\documentclass{article} %
\usepackage{fullpage}
\usepackage{authblk}
\usepackage{natbib}

\usepackage{color}
\usepackage{xcolor}
\usepackage{hyperref}
\hypersetup{
    colorlinks,
    linkcolor={red!50!black},
    citecolor={blue!50!black},
    urlcolor={blue!80!black}
}
\usepackage{enumerate}
\usepackage{mathrsfs}
\usepackage{mathtools}
\usepackage[font=small,labelfont=bf]{caption}

\usepackage{amsmath,amsthm,amssymb,colonequals,etoolbox}
\usepackage{thmtools}
\usepackage{url}
\usepackage{cleveref}
\usepackage{microtype}
\usepackage{thm-restate}

\bibpunct{(}{)}{;}{a}{,}{,}

\renewcommand{\geq}{\geqslant}

\renewcommand{\leq}{\leqslant}

\newtheorem{thm}{Theorem}[section]

\newtheorem{myprop}[thm]{Proposition}
\newtheorem{mylemma}[thm]{Lemma}

\newtheorem{myfact}[thm]{Fact}

\newtheorem{assmp}[thm]{Assumption}

\DeclareMathOperator*{\argmin}{arg\,min}

\DeclareMathOperator*{\Tr}{\mathrm{tr}}

\newcommand{\calV}{\mathcal{V}}

\newcommand{\calL}{\mathcal{L}}

\newcommand{\R}{\ensuremath{\mathbb{R}}}

\newcommand{\N}{\ensuremath{\mathbb{N}}}

\newcommand{\norm}[1]{\lVert #1 \rVert}
\newcommand{\bignorm}[1]{\left\lVert #1 \right\rVert}

\newcommand{\bigip}[2]{\ensuremath{\left\langle #1, #2 \right\rangle}}
\newcommand{\ip}[2]{\ensuremath{\langle #1, #2 \rangle}}

\newcommand{\E}{\mathbb{E}}
\newcommand{\abs}[1]{\ensuremath{| #1 |}}
\newcommand{\bigabs}[1]{\ensuremath{\left| #1 \right|}}

\newcommand{\ind}{\mathbf{1}}

\renewcommand{\Pr}{\mathbb{P}}
\newcommand{\T}{\mathsf{T}}

\newcommand{\calD}{\mathcal{D}}

\newcommand{\calE}{\mathcal{E}}

\newcommand{\calG}{\mathcal{G}}
\newcommand{\calF}{\mathcal{F}}

\newcommand{\calR}{\mathcal{R}}
\newcommand{\calH}{\mathcal{H}}

\newcommand{\calZ}{\mathcal{Z}}
\newcommand{\calP}{\mathcal{P}}
\newcommand{\scrF}{\mathscr{F}}

\numberwithin{equation}{section}

\newcommand{\tvnorm}[1]{\norm{#1}_{\mathrm{tv}}}

\newcommand{\e}{\varepsilon}
\newcommand{\rmd}{\mathrm{d}}

\newcommand{\sfX}{\mathsf{X}}

\newcommand{\bbS}{\mathbb{S}}

\DeclarePairedDelimiterX{\infdivx}[2]{(}{)}{%
  #1\;\delimsize\|\;#2%
}
\newcommand{\KL}{\mathrm{KL}\infdivx}

\newcommand{\Rlin}{R_{\mathrm{lin}}}
\newcommand{\Rind}{R_{\mathrm{ind}}}

\title{Shallow diffusion networks provably learn hidden low-dimensional structure}

\author[1]{Nicholas M. Boffi}
\author[2]{Arthur Jacot}
\author[3]{Stephen Tu}
\author[4]{Ingvar Ziemann}
\affil[1]{Carnegie Mellon University}
\affil[2]{Courant Institute of Mathematical Sciences, New York University}
\affil[3]{Department of Electrical and Computer Engineering, University of Southern California}
\affil[4]{Department of Electrical and Systems Engineering, University of Pennsylvania}

\begin{document}

\maketitle

\begin{abstract}
Diffusion-based generative models provide a powerful framework for learning to sample from a complex target distribution. 
The remarkable empirical success of these models applied to high-dimensional signals, including images and video, stands in stark contrast to classical results highlighting the curse of dimensionality for distribution recovery.
In this work, we take a step towards understanding this gap through a careful analysis of learning diffusion models over the Barron space of single layer neural networks.
In particular, we show that these shallow models provably adapt to simple forms of low dimensional structure, thereby avoiding the curse of dimensionality.
We combine our results with recent analyses of sampling with diffusion models to provide an end-to-end sample complexity bound for learning to sample from structured distributions. 
Importantly, our results do not require specialized architectures tailored to particular latent structures, and instead rely on the low-index structure of the Barron space to adapt to the underlying distribution.

\end{abstract}

\section{Introduction}
\label{sec:intro}
Generative models learn to sample from a target probability distribution given a dataset of examples.
Applications are pervasive, and include language modeling~\citep{li2022diffusionLM}, high-fidelity image generation~\citep{rombach2022latentdiffusion}, \textit{de-novo} drug design~\citep{watson2023novo}, and molecular dynamics~\citep{arts2023moleculardynamics}.
Recent years have witnessed extremely rapid advancements in the field of generative modeling, particularly with the development of models based on dynamical transport of measure~\citep{santambrogio2015optimal}, such as diffusion-based generative models~\citep{ho2020ddpm,song2021scorebased}, stochastic interpolants~\citep{albergo2023stochastic}, flow matching~\citep{lipman2023flow}, and rectified flow~\citep{liu2022}~approaches.
Yet, despite their strong empirical performance and well-grounded mathematical formulation, a theoretical understanding of \textit{how} and \textit{why} these large-scale generative models work is still in its infancy.

A promising line of recent research has shown that the problem of sampling from an arbitrarily complex distribution can be reduced to unsupervised learning: for diffusion models, if an accurate velocity or score field can be estimated from data, then high-quality samples can be generated via numerical simulation~\citep{chen2023sampling,lee2023convergence}.
While deeply insightful, these works leave open the difficulty of statistical estimation, and therefore raise the possibility that the sampling problem's true difficulty is hidden in the complexity of learning.

In this work, we address this fundamental challenge by presenting an end-to-end analysis of sampling with score-based diffusion models.
To balance tractability of the analysis with empirical relevance, we study the Barron space of single-layer neural networks~\citep{weinan_2019_barron,bach2017breaking}.
This space contains important features of models used in practice -- most importantly, parametric nonlinearity -- while retaining well-studied theoretical properties that we can adapt to the generative modeling problem.
As a paradigmatic example of the widely-held belief that real-world datasets contain hidden low-dimensional structure~\citep{tenenbaum2000isomap,weinberger2006unsupervised}, we focus on an idealized setting in which the target data density is concentrated on an unknown low-dimensional linear manifold.
We show that for learning to sample from a target distribution supported on a low-dimensional subspace, 
diffusion models backed by single layer neural networks -- which we refer to as \emph{shallow diffusion networks} -- enjoy a sample complexity bound that only depends exponentially on the dimension of the subspace rather than on the ambient dimension. 
In addition, we extend our framework to the setting of target distributions constructed by composing independent components. 
Our results highlight that diffusion models based on shallow neural networks without specific architectural modifications can adapt to hidden structure and sidestep the curse of dimensionality; in this way, they give insight into the empirical performance of more complex network architectures on real-world high-dimensional datasets.

\section{Related Work}
\label{sec:related_work}

\paragraph{Sampling bounds for diffusion models.}
Many recent analyses of diffusion models have focused on 
the accuracy of sampling from a discretized process assuming 
access to
an $L_2$ accurate score function.
In this direction, both discretized SDEs~\citep{lee2022polynomial,chen2023sampling,lee2023convergence,chen2023improved,benton2024nearly} and probability flow ODEs~\citep{chen2023ddim,chen2023probabilityflow,li2024sharp,liang2024non,gao2024convergence} have been studied.
Recent work by \citet{li2024adapting} shows that the DDPM sampler~\citep{ho2020ddpm} can be tuned so that in the presence of low-dimensional structure, the discretization error only depends polynomially on the intrinsic dimension (in addition to the score error). 
However, because these works assume the existence of an $\e$-accurate score function in $L_2$, they leave open the question of the sample complexity of obtaining such a model, which is precisely what we tackle here.

\paragraph{Sample complexity of score matching.}
\citet{block2020generative} and~\citet{koehler2023statistical} employ the standard Rademacher complexity framework to bound the error of empirical risk minimization for learning a score function with the implicit score matching objective, but leave open the question of which function class to learn over.
\citet{han2024neural} and~\citet{wang2024evaluating} consider optimizing the denoising score matching loss over neural network models, and show that gradient descent on overparameterized models finds high-quality solutions.
\citet{wibisono2024optimal},~\citet{zhang2024minimax},~\citet{oko2023diffusion}, and~\citet{dou2024optimal} study the minimax optimality of diffusion models for distribution estimation under various functional assumptions on the target density and its score. 
Taken together, these works show that diffusion modeling is both statistically nearly-optimal and computationally efficient for learning to sample.
Yet, simultaneously, they highlight the presence of the curse of dimensionality in the absence of structured data.

To address this issue, 
both~\citet{oko2023diffusion} and~\citet{chen2023subspace}
study a setting in which the data lives on a low-dimensional subspace, and show that this latent structure avoids exponential dependencies on the ambient dimension.
However, both works require assumptions about the low-dimensional subspace (i.e., knowledge of the dimension) and/or constraints on the network architecture (i.e., bounded weight sparsity) which are usually not available in practice and/or computationally challenging to implement.
Our work can be seen as further bridging the gap between theory and practice in this setting by showing that the same dependence on the latent dimension also holds for shallow Barron networks which
are closer to the architectures used in practice; we defer a detailed comparison to \Cref{sec:results}.
Concurrent with our work, \citet{azangulov2024convergence} 
show that the network architecture studied in 
\citet{oko2023diffusion} can also be used to learn diffusion models 
for data residing on general compact smooth manifolds.
Earlier work from \citet{debortoli2022convergence} also studies a similar setting as \citet{azangulov2024convergence},
and proves a bound that 
depends exponentially on the diameter of the manifold.
Finally,~\citet{cole2024scorebased} show that under the assumption that the target log-relative density (w.r.t.\ a standard Gaussian) can be approximated by a NN with low path norm, score estimation can be performed with a sample complexity bound that does not depend explicitly on the ambient dimension.\footnote{However, there are still $O(1)^D$ pre-factors in the final rate.} However their absolute continuity assumption rules out examples such as target distributions supported on low dimensional manifolds.

\paragraph{Learning in Barron spaces.} 
Even though the implicit bias of DNNs remains a largely open question, there is now strong consensus that the implicit bias of shallow networks with large width is accurately captured by the so-called Barron norm \citep{weinan_2019_barron} or the (total variation) $\calF_1$ norm \citep{bach2017breaking}. 
These networks have also been analyzed using statistical physics-based techniques in the mean-field limit~\citep{rotskoff_trainability_2019, mei_mean_2018, sirignano_mean_2020}.
Such networks can avoid the curse of dimensionality when the target function has \emph{low-index} structure, i.e., when $f(x)=f(Px)$ for a low-dimensional projection $P$, leading to generalization bounds that depend on the intrinsic dimension $d$ of $Px$ rather than the ambient dimension $D$ of $x$~\citep{bach2017breaking}. 
A number of recent results have also studied the (sometimes modified) gradient descent dynamics of shallow networks, and how this low-index structure emerges in the network \citep{abbe2022merged_staircase,bietti_2022_single_index,ben_2022_effective_dyn,glasgow2024_sgd_xor_shallow,lee_2024_low_polynomial}. While most of this literature focuses on supervised training problems, some work has shown that this type of analysis can be extended to the unsupervised case, in particular to learn energy-based models \citep{domingo_2021_F1_energy}.

\section{Problem Formulation and Main Results}
\label{sec:results}
Our goal in this work is to study the statistical complexity of learning to sample from a target probability measure $p_0(x_0)$ defined on $\R^D$ given a dataset of $n$ iid examples $x_0^i \sim p_0$ for $i = 1, \hdots, n$. 
In particular, we consider the use of a diffusion model~\citep{sohldickstein2015unsupervised,song2021scorebased,ho2020ddpm} to learn a stochastic process that maps random noise to a new sample from the data distribution.
We assume that the target $p_0$ contains hidden latent structure -- either a low-dimensional subspace or independent components -- and our primary goal will be to show that a shallow network can learn this hidden structure efficiently, in the sense that the statistical rates are governed primarily by the underlying latent dimension $d$, as opposed to the ambient dimension $D \gg d$.

\paragraph{Diffusion models.} 
We consider diffusion-based generative models based on stochastic differential equations~\citep{song2021scorebased}.
These models construct a path in the space of measures between the target $p_0$ and a standard Gaussian $\mathsf N(0, I_D)$ by defining a simple \textit{forward process} that converges to Gaussian data over an infinite horizon.
For simplicity, we study the simple Ornstein-Uhlenbeck (OU) process,
\begin{align}
    \label{eqn:ou}
    \rmd x_t = - x_t \rmd t + \sqrt{2} \rmd B_t, \quad t \in [0, T],
\end{align}
though our results straightforwardly generalize to the time-scaled OU processes commonly used in practice~\citep{song2021scorebased}. 
In~\eqref{eqn:ou}, $(B_t)_{t \geq 0}$ denotes a standard Brownian motion on $\R^D$.
Due to the linear nature of the OU process, for $w \sim \mathsf N(0, I_D)$ drawn independently of $x_0$, $x_t$ is equivalent in distribution to the stochastic interpolant~\citep{albergo2023stochastic}
\begin{align}
    x_t \overset{\mathsf d}{=} m_t x_0 + \sigma_t w, \quad m_t := \exp(-t), \quad \sigma_t := \sqrt{1 - \exp(-2t)}.
\end{align}
Let the marginal distributions of $x_t$ be denoted as $(p_t)_{t \in [0, T]}$.
The \emph{reverse process} is the process of $y_t := x_{T-t}$ for $t \in [0, T]$. 
A classic result~\citep{anderson1982diffusion} shows that the reverse process satisfies
\begin{align}
    \rmd y_t = ( y_t + 2 \nabla \log p_{T-t}(y_t) )\rmd t + \sqrt{2} \rmd B_t, \quad y_0 \sim p_T(\cdot), \quad t \in [0, T]. \label{eq:diffusion_reverse_process}
\end{align}
Thus, assuming knowledge of the time-dependent score function $\nabla \log p_t$, sampling from $p_0(\cdot)$ can be accomplished by (a) setting $T$ large enough so that
$p_T(\cdot)$ is approximately an isotropic Gaussian, (b) sampling $y_0 \sim \mathsf N(0, I_D)$,
and (c) running the reverse SDE \eqref{eq:diffusion_reverse_process} until time $T$.

To implement this scheme in practice,
the score function must be learned, and the reverse process must be discretized.
Assuming access to a learned score
function $\hat{s} \approx \nabla \log p$, we now consider
discretizing \eqref{eq:diffusion_reverse_process}.
In this work we make use of the \emph{exponential integrator} (EI),
which fixes a sequence (to be specified) of reverse process timesteps
$0 = \tau_0 < \tau_1 < \dots < \tau_N = T$
and implements
\begin{align}
    \rmd \tilde{y}_t = ( \tilde{y}_t + 2 \nabla \log \hat{s}_{T - \tau_k}(\tilde{y}_{\tau_k}) ) \rmd t + \sqrt{2} \rmd B_t, \quad t \in [\tau_k, \tau_{k+1}], \quad k \in \{0, \dots, N-1\}. \label{eq:exponential_integrator}
\end{align}
Recently, building off of the works by~\citet{chen2023sampling} and~\citet{lee2023convergence},~\cite{benton2024nearly} showed that it suffices to control the score approximation error in $L_2(p_t)$ to guarantee that 
the process~\eqref{eq:exponential_integrator} yields a high quality sample from $p_0$.\footnote{Technically,
\cite{benton2024nearly} 
guarantees a high quality sample from $p_{T - \tau_{N-1}}(\cdot)$ instead of $p_0(\cdot)$.}

\paragraph{Score function estimation.}
To estimate the score function $\nabla \log p_t$ over the interval $[0, T]$, one would ideally minimize the least-squares objective over a model $\hat{s}$,
\begin{align}
    \calR(\hat{s}) := \int_0^T \calR_t(\hat{s}_t) \,\rmd t,
    \qquad \calR_t(\hat{s}_t) := \E_{x_t} \norm{\hat{s}_t(x_t) - \nabla \log p_t(x_t) }^2 . 
    \label{eq:score_loss}
\end{align}
While direct minimization is not possible because $\nabla\log p_t$ is not observed, minimizing $\calR(s)$ is equivalent (up to a constant) to minimizing the following 
denoising score matching (DSM) loss~\citep{vincent2011denoising}
\begin{align}
    \calL(\hat{s}) := \int_0^T\calL_t(\hat{s}_t) \,\rmd t,
    \qquad \calL_t(\hat{s}_t) := \E_{(w, x_t)} \norm{\hat{s}_t(x_t) + w/\sigma_t }^2, \label{eq:dsm_loss}
\end{align}
as can be shown by observing that Tweedie's identity~\citep{efron2011tweedie} implies \begin{equation}
    \nabla\log p_t(x) = -\frac{1}{\sigma_t}\E[w \mid x_t = x].
    \label{eqn:tweedie}
\end{equation}
In practice,~\eqref{eq:dsm_loss} is typically approximated via Monte-Carlo by generating samples $x_{t^i}^i = m_{t^i}x_0^i + \sigma_{t^i}w^i$ using the dataset of samples from $p_0$, iid random draws of Gaussian noise $w^i$, and random time points $t^i$ drawn from $[0, T]$.
This empirical risk can then be minimized to estimate a time-dependent score function $\hat{s}: [0, T] \times \R^D \rightarrow \R^D$.

In this work, to simplify the mathematical analysis, we consider a stylized variant in which we fix
a sequence of timesteps
$0 < t_0 < \dots < t_{N-1} = T$ 
and estimate $N$ time-independent score functions $\{\hat{s}_{t_i}\}_{i=0}^{N-1}$ of the form $\hat{s}_{t_i} : \R^D \rightarrow \R^D$.
\begin{align}
    \hat{s}_{t} \in \argmin_{s_{t} \in \scrF_{t}} \hat{\calL}_{t}(s_{t}), \quad 
    \hat{\calL}_{t}(s_t) := \frac{1}{n}\sum_{i=1}^{n} \norm{s_t(x_t^i) + w^i/\sigma_{t^i}}^2, %
    \quad t \in \{t_i\}_{i=0}^{N-1}.
    \label{eq:dsm_empirical_loss_single_scale}
\end{align}
In the sequel, we will let $\calD_t := \{(x_0^i, x_t^i)\}_{i=1}^{n}$ denote the training data used in \eqref{eq:dsm_empirical_loss_single_scale}
for timestep $t$.

From \citet{wibisono2024optimal}, 
we know that if the true score $\nabla \log p_t$ is Lipschitz continuous and $p_t$ is sub-Gaussian, 
then the minimax rate for estimating $\nabla \log p_t$ is
given by $n^{-2/(D+4)}$.\footnote{Interestingly, this rate is slower than the $n^{-2/(D+2)}$ rate for learning Lipschitz functions~\citep{tsybakov2008book}.}
Unfortunately, this type of bound ignores all latent structure, raising the question of whether or not diffusion models can learn latent structure in a sample efficient way.

\paragraph{The Barron space $\calF_1$.} 
Consider a shallow neural network $f_m$ with $m$ neurons and mean-field scaling, $f_m(x) = \frac{1}{m}\sum_{i=1}^m u_i\sigma(\ip{x}{v_i})$.
In the limit as $m$ tends to infinity, the summation may be replaced by integration $f_\infty(x) = \int u\sigma(\ip{x}{v})\, \rmd\mu(u,v)$ against a signed Radon measure $\mu$ over the neuron parameters $(u,v)$. 
This leads to the Barron space $\calF_1$~\citep{bach2017breaking,mhaskar2004neuralnetworks,rotskoff_trainability_2019,mei_mean_2018,sirignano_mean_2020}, which models shallow neural networks in the infinite width limit and in the feature learning regime~\citep{chizat_lazy_2020}.

Concretely, given a Radon measure $\mu$ on a measurable space $\calV$, recall that the \emph{total variation norm} (TV) is defined as $\tvnorm{\mu} := \sup_{g : \calV \mapsto [-1, 1], \,\textrm{$g$ cts}} \int_{\calV} g(v) \, \rmd\mu(v)$.
Given a basis function $\varphi_v(x)$, the TV-norm induces the space of functions $\calF_1 := \{ f(x) = \int_{\calV} \varphi_v(x) \,\rmd\mu(v) \mid \tvnorm{\mu} < \infty \}$.
For a function $f \in \calF_1$, its $\calF_1$-norm is the infimum over all TV-norms of measures that can represent $f$, i.e., $\norm{f}_{\calF_1} := \inf\{ \tvnorm{\mu} \mid f(x) = \int_{\calV} \varphi_v(x) \,\rmd\mu(v) \}$.

In this work, we will consider the special case $\calV = \bbS^{p-1} \times \bbS^{p-1}$ for $p \in \N_+$ and $\varphi_v(x) = u \sigma(\ip{x}{v})$, where $(u, v) \in \calV$ and $\sigma(\cdot) = \max\{0, \cdot\}$ is the ReLU activation.
Hence, the induced class $\calF_1$ consists of vector-valued maps from $\R^p \to \R^p$, 
which satisfy $\norm{f(x)} \leq \norm{f}_{\calF_1} \norm{x}$.
In learning the score functions $\hat{s}_{t_i}$ 
via DSM~\eqref{eq:dsm_empirical_loss_single_scale}, we will utilize norm-ball subsets of $\calF_1$ to model the true score functions $\nabla \log p_{t_i}(x)$.
This allows us to leverage the low-index structure of $\calF_1$~\citep{bach2017breaking} and obtain bounds for score estimation that scale with the \emph{intrisic}, rather than ambient, dimension of the problem. 

\paragraph{Notation:}
We briefly review the (relatively standard) notation used in this work.
For a $d$-dimensional vector $x$, the $\ell_p$ norm is denoted $\norm{x}_p$; the notation $\norm{x}$ is reserved for the Euclidean ($p=2$) case.
The (closed) $\ell_2$-ball of radius $r$ in $d$-dimension is denoted by $B_2(r, d)$; when $r=1$, we use the shorthand $B_2(d)$.
The unit sphere in $\R^d$ is denoted $\bbS^{d-1}$.
The notation $O_d(\cdot)$ hides both universal constants and constants that depend arbitrarily on the variable $d$.
Similarly, $\tilde{O}_d(\cdot)$ 
hides universal constants, constants that depend on $d$, and terms that may depend poly-logarithmically on $d$, i.e., terms of the form $\log^{O(d)}(\cdot)$.
The notation $\mathrm{poly}(\cdot)$ indicates a polynomial dependence on the arguments, whereas $\mathrm{poly}_d(\cdot)$ indicates that the polynomial degrees are allowed to depend arbitrarily on $d$.
Finally, the notation $a \lesssim b$ (resp.~$a \gtrsim b$) indicates that there exists a universal positive constant $c$ such that $a \leq c b$ (resp.~$a \geq c b$).

\subsection{Learning latent subspace structure}
\label{sec:results:subspace}
We first consider a setting in which $x_0$ is supported on a $d$-dimensional ($d \ll D$) linear subspace.
Specifically, we study
\begin{align}
    x_0 = U z_0, \quad z_0 \sim \pi_0(\cdot), \quad U \in O(D, d), \label{eq:subspace_structure}
\end{align}
where $z_0$ is a $d$-dimensional random vector
and $O(D, d) := \{ U \in \R^{D \times d} \mid U^\T U = I_d\}$ denotes the $d$-dimensional orthogonal group in $\R^D$.
Note that both the subspace dimension $d$ and the embedding matrix $U$ are unknown to the learner.

Recently, both \citet{chen2023subspace} and \citet{oko2023diffusion} 
consider learning diffusion models under the subspace structure
\eqref{eq:subspace_structure}.
The main takeaway from both works is that the latent subspace dimension $d$, rather than the ambient dimension $D$, can govern the complexity of learning to sample from $p_0$ if the network used for learning satisfies various architectural assumptions.
While insightful, these architectural assumption are difficult to satisfy in practice.
\citet{chen2023subspace}
utilize a function class that is specifically tailored to the linear structure~\eqref{eq:subspace_structure}, in the sense that a linear autoencoder with prior knowledge of the latent dimension $d$ is used to reduce the learning problem to the latent space.
The situation is improved in~\citet{oko2023diffusion}, which considers fully-connected neural networks with bounded weight \emph{sparsity}.
Though closer to real-world architectures, 
optimizing networks with bounded sparsity constraints is computationally challenging in practice.

Our first result further closes the gap between theory and practice: we show that by optimizing over $\calF_1$, the space of infinite-width shallow networks, the latent subspace is adaptively learned without requiring prior latent dimension knowledge or difficult to impose sparsity constraints.
We note that from a computational perspective, sufficiently wide shallow networks trained with gradient descent (GD) and weight decay will converge to the minimal $\calF_1$-norm solution \citep{Chizat2018}.
While control on the number of neurons required in the worst case is not
available (optimizing over $\calF_1$ is unfortunately NP-hard~\citep{bach2017breaking}),
recent results have shown that 
under various assumptions on the data and task, this hardness can be avoided and GD can indeed learn low-index functions with a number of neurons that scales polynomially with an exponent that depends only on the intrinsic dimension~\citep{abbe2022merged_staircase,dandi_2023_giant_steps_low_index, lee_2024_low_polynomial}.

Towards stating our first result, we begin with a few standard regularity assumptions.
\begin{assmp}
\label{assumption:sub_gaussian_z_0}
The latent variable $z_0 \sim \pi_0(\cdot)$ is $\beta$-sub-Gaussian\footnote{This definition is also referred to as \emph{norm-}sub-Gaussian in the literature~\citep[see e.g.][]{jin2019short}.} (with $\beta \geq 1$),
i.e.,
$$
    \E \exp(\lambda (\norm{z_0}-\E \norm{z_0})) \leq \exp(\lambda^2 \beta^2/2), \quad \forall \lambda \in \R.
$$
\end{assmp}
\noindent
Our next assumption concerns the regularity of the
score function of the latent distribution $\pi_t$, which denotes the marginal distribution of $z_t := U^\T x_t$.
\begin{assmp}
\label{assumption:lipschitz_pi_t}
The latent score $\nabla \log \pi_t$ is $L$-Lipschitz (with $L \geq 1$) on $\R^d$ for all $t \geq 0$.
\end{assmp}
\noindent
We emphasize that \Cref{assumption:lipschitz_pi_t} concerns the Lipschitz regularity of the latent measure $\pi_t$ and \emph{not} the ambient measure $p_t$; due to the subspace structure \eqref{eq:subspace_structure}, the Lipschitz constant of the ambient score diverges as $t \to 0$.
Finally, we define some shorthand notation for $t \in [0, T]$,
\begin{align*}
    \mu_{t,x} := \mu_0 \vee \sigma_t\sqrt{D}, \quad \mu_{t,z} := \mu_0 \vee \sigma_t\sqrt{d}, \quad \bar{L} := c L(\mu_{0} + \sqrt{d} + \norm{\nabla \log \pi_0(0)}),
\end{align*}
where $\mu_0 := (\E \norm{z_0}^2)^{1/2}$ and $c \geq 1$ is a universal constant.
Since $\sigma_t \to 1$ as $t \to \infty$, 
these constants are uniformly bounded above. 
Hence, we define 
$\mu_{x} := \lim_{t\rightarrow\infty}\mu_{t,x}$ and 
$\mu_{z} := \lim_{t\rightarrow\infty}\mu_{t,z}$.
\begin{restatable}{mythm}{scoreestimationlatent}
\label{thm:score_estimation_latent}
Suppose that $p_0$ follows the latent structure
\eqref{eq:subspace_structure}, and that both \Cref{assumption:sub_gaussian_z_0} and \Cref{assumption:lipschitz_pi_t} hold.
Fix a $t > 0$ and define 
\begin{align}
    \scrF_t := \{ s : \R^D \mapsto \R^D \mid \norm{s}_{\calF_1} \leq R_t \}, \quad R_t := \bar{R}_t n^{\frac{d+1}{2(d+5)}} + \frac{D}{\sigma_t^2}, \label{eq:F_t_latent}
\end{align}
where $\bar{R}_t$ does not depend on $n$.\footnote{The explicit dependence of $\bar{R}_t$ on the other problem parameters is detailed in the proof.}
Suppose that $n$ satisfies 
\begin{align}
    n \geq n_0(t) := \mathrm{poly}(D, 1/\sigma_t, \mu_{t,x} \vee \beta) \cdot \mathrm{poly}_d(\bar{L}, \mu_{t,z} \vee \beta). \label{eq:ERM_score_matching_subspace_burnin}
\end{align}
Then, the empirical risk minimizer $\hat{s}_t \in \argmin_{s \in \scrF_t} \hat{\calL}_t(s)$ satisfies:
\begin{align}
    \E_{\calD_t}[\calR_t(\hat{s}_t)] \leq \tilde{O}_d(1) \left[  \frac{D^2}{\sigma_t^2 n}(\bar{L} (\mu_{t,z} \vee \beta))^{d+3} (\mu_{t,x} \vee \beta)^2 \right]^{\frac{2}{d+5}} + \tilde{O}_d(1)  \sqrt{ \frac{D^3}{\sigma_t^6 n}(\mu_{t,x} \vee \beta)^2 }. \label{eq:ERM_score_matching_subspace_bound}
\end{align}
\end{restatable}
Some remarks regarding \Cref{thm:score_estimation_latent}
are in order.
First, we can upgrade \Cref{thm:score_estimation_latent} to a high-probability bound with minor modifications to the proof; we omit these
details in the interest of brevity.
Second, we note that our $n^{2/(d+5)}$ rate nearly matches the minimax optimal rate of score matching from~\citet{wibisono2024optimal}, but with the subspace dimension $d$ replacing the ambient dimension $D$; we leave showing a $n^{2/(d+4)}$ rate to future work.
Last, as noted before,~\citet[Theorem 2]{chen2023subspace} proves a related result for learning score functions
under the subspace structure \eqref{eq:subspace_structure}.
However, our result in~\Cref{thm:score_estimation_latent} substantially
improves their result in the following ways.
First, as already mentioned, our result does not require any specialized architectures, but instead applies to learning directly in the Barron space $\calF_1$ of shallow neural networks.
Additionally, our result also provides several technical improvements:
(a) our leading dependence on $n$ is improved to $n^{2/(d+5)}$ from
$n^{(2-o(1))/(d+5)}$,\footnote{However, their logarithmic dependence on $n$ is only through $\log^{O(1)}(n)$ terms instead of our $\log^{O(d)}(n)$.}
and (b) our dependence on $\sigma_t$ is improved to
$\sigma_t^{-4/(d+5)}$ instead of $\sigma_t^{-2}$; as $t \to 0$ the former  degrades slower than the latter.

We now use \Cref{thm:score_estimation_latent}
to provide an end-to-end sample complexity bound for sampling from $p_0$.
\begin{restatable}{mycorollary}{samplingsubspace}
\label{corollary:sampling_subspace}
Fix $\e, \zeta \in (0, 1)$.
Suppose that $p_0$ follows the latent structure
\eqref{eq:subspace_structure}, and that both \Cref{assumption:sub_gaussian_z_0} and \Cref{assumption:lipschitz_pi_t} hold.
Consider the exponential integrator \eqref{eq:exponential_integrator} with:
\begin{align}
    T = c_0 \log\left(\frac{\sqrt{D} \vee \mu_0}{\e}\right), \quad N = 2 \left\lceil c_1 \frac{D \vee \mu_0^2}{\e^2} \left[  \log^2\left(\frac{\sqrt{D} \vee \mu_0}{\e}\right) + \log^2\left(\frac{1}{\zeta}\right) \right] \right\rceil, \label{eq:diffusion_T_N_setting}
\end{align}
and reverse process discretization timesteps $\{\tau_i\}_{i=0}^{N}$ defined as:
\begin{align}
    \tau_i = \begin{cases} 
        2(T-1)\frac{i}{N} &\text{if } i \in \{0, \dots, N/2\}, \\
        T - \zeta^{2i/N-1} &\text{if } i \in \{N/2+1,\dots,N\}.
        \end{cases}
    \label{eq:reverse_process_timesteps}
\end{align}
Next, define the forward process
timesteps $\{t_i\}_{i=0}^{N-1}$ by $t_i := T - \tau_{N-i}$.
Suppose the exponential integration scheme is run with score functions $\{ \hat{s}_{t_i} \}_{i=0}^{N-1}$,
where $\hat{s}_{t_i} \in \argmin_{s \in \scrF_{t_i}} \hat{\calL}_{t_i}(s)$ with
$\scrF_{t}$ as defined in \eqref{eq:F_t_latent}.
Suppose furthermore that $n$ satisfies:
\begin{align*}
    n \geq \tilde{O}_d(1) \max\left\{  \frac{D^2}{\zeta}(\bar{L} (\mu_{z} \vee \beta))^{d+3} (\mu_{x} \vee \beta)^2 \cdot \e^{-(d+5)}, \frac{D^3}{\zeta^3} (\mu_x \vee \beta)^2 \cdot \e^{-4}, n_0(\zeta) \right\},
\end{align*}
where $n_0(\cdot)$ is defined in \eqref{eq:ERM_score_matching_subspace_burnin}.
With constant probability (over the randomness of the training datasets $\{\calD_{t_i}\}_{i=0}^{N-1}$), we have that
$\KL{p_\zeta}{\mathrm{Law}(\hat{y}_{T-\zeta})} \leq \e^2$.    
\end{restatable}

Treating $\bar{L}$ and $\beta$ as constants,
\Cref{corollary:sampling_subspace} prescribes a rate of $n \geq \tilde{O}_d(1) \frac{\mathrm{poly}(D)}{\zeta} \e^{-(d+5)}$ (after a burn-in on $n$) to obtain a sampler that satisfies $\KL{p_\zeta}{\mathrm{Law}(\hat{y}_{T-\zeta})} \leq \e^2$. 
To the best of our knowledge, this is the first end-to-end sample complexity bound for learning a diffusion model over shallow neural networks that adapts to the intrinsic dimensionality of the problem.
Note that as \Cref{corollary:sampling_subspace} controls the 
KL-divergence between the true data distribution $p_\zeta$ 
and the distribution $\mathrm{Law}(\hat{y}_{T-\zeta})$
of the final iterate of the exponential integrator \eqref{eq:exponential_integrator},
by Pinsker's inequality this also implies control on the TV-distance
$\tvnorm{ p_\zeta - \mathrm{Law}(\hat{y}_{T-\zeta}) }$.
Furthermore, we can upgrade \Cref{corollary:sampling_subspace}
to a high probability guarantee by utilizing a high probability variant of \Cref{thm:score_estimation_latent}.

The parameter $\zeta > 0$ is the early stopping parameter which is found
in practical implementations of diffusion models (cf.~\citet{karras2022edm}).
Note that this is necessary since $p_0$ is supported on a lower-dimensional manifold and hence
$\nabla \log p_0$ is not smooth on all of $\R^D$.
We remark that bounds comparing the original $p_0$ to $\mathrm{Law}(\hat{y}_{T-\zeta})$ are possible in Wasserstein distance
by adopting the techniques from e.g.~\citet[Section 3.2]{chen2023sampling}; we omit these calculations.

Compared to \citet[Theorem 3]{chen2023subspace}, who obtain a
$n \geq \tilde{O}_d(1) (\e \sqrt{\zeta})^{-(d+5)/(1-o(1))}$ rate 
in the case of a latent subspace-aware architecture, we see that our bound also improves the dependency on $\zeta$.
This is important, because ultimately $\zeta$ will be chosen to decay to zero as $n \to \infty$.
We do remark, however, that \citet[Theorem 3]{chen2023subspace} only 
depends polylogarithmically on the ambient dimension $D$ instead of polynomially. This can be traced back in their analysis to imposing the constraint that their score functions are uniformly bounded,
i.e., $\sup_{z, t} \norm{s(z, t)} \leq K$, which allows truncation arguments to avoid picking up extra $\mathrm{poly}(D)$ factors.
We choose to not impose such constraints in our model class, as this adds another hyperparameter that must be tuned in practice. 
We leave open the question of whether or not these $\mathrm{poly}(D)$ pre-factors in the sample complexity can be removed without further modifications (e.g., clipping) of the $\calF_1$ hypothesis class.

Compared to \citet[Theorem 6.4]{oko2023diffusion}, 
\Cref{corollary:sampling_subspace} also relaxes a few technical assumptions, 
including a uniformly lower bounded density $\pi_0$, and a requirement that $\pi_0$ be $C^\infty$ near the boundary $[-1, 1]^d$.
On the other hand, their work obtains
a sharper rate on $W_1(p_0, \mathrm{Law}(\hat{y}_{T-\zeta})) \lesssim n^{-(3-\delta)/(d+4)}$ for any $\delta > 0$.
We also leave open the question of whether these extra assumptions can be used to strengthen our guarantees for learning over $\calF_1$.

\subsection{Extensions to independent component structure}
\label{sec:results:independent_components}
We now consider a different type of latent structure -- here generated by independence -- as opposed to the low-dimensional subspace setting just studied.
Specifically, we suppose that for some $K \in [D]$,
\begin{align}
    x_0 = U z_0, \quad z_0 \sim (z_0^{(1)}, \dots, z_0^{(K)}), \quad z_0^{(i)} \sim \pi_0^{(i)}(\cdot), \label{eq:independent_components}
\end{align}
where $U \in O(D) := \{ U \in \R^{D \times D} \mid U^\T U = I_D \}$,
$z_0^{(i)} \in \R^{d_i}$ with $\sum_{i=1}^{K} d_i = D$, 
and where $z_0^{(i)}$ is independent of $z_0^{(j)}$ for $i \neq j$.
Similar to the linear subspace setting, we assume that the orthonormal matrix $U$, the number of components $K$, and the dimensionality of each component $\{d_i\}_{i=1}^K$ are all unknown to the learner, and we study the adaptive properties of $\calF_1$ in the presence of this latent structure.
We begin by imposing a similar set of assumptions as in the subspace case (cf.~\Cref{sec:results:subspace}).

\begin{assmp}
\label{assumption:sub_gaussian_z_0_components}
For all $i \in [K]$, we have that $\pi_0^{(i)}$ is $\beta_i$-sub-Gaussian (for $\beta_i \geq 1$), i.e.,
$$
    \E \exp(\lambda(\norm{z_0^{(i)}} - \E\norm{z_0^{(i)}})) \leq \exp(\lambda^2 \beta_i^2 / 2), \quad \lambda\in\R.
$$
\end{assmp}

\noindent
Our next assumption again deals with the latent measure $\pi_t$, defined as the marginal distribution of $z_t := U^\T x_t$.
We decompose $z_t = (z_t^{(1)}, \dots, z_t^{(K)})$ into coordinate groups as for $z_0$, and we define $\pi_t^{(i)}$ as the marginal distribution of $z_t^{(i)}$.

\begin{assmp}
\label{assumption:lipschitz_score_functions}
For all $i \in [K]$,
$\nabla \log \pi_t^{(i)}$ is $L_i$-Lipschitz (for $L_i \geq 1$) on $\R^{d_i}$ for all $t \geq 0$.
\end{assmp}

\noindent
As before, we define some shorthand notation for $t \in [0, T]$ and $i \in [K]$:
\begin{align*}
    \mu_{t,x}^{(i)} := \mu_0^{(i)} \vee \sigma_t \sqrt{d_i}, \quad \bar{L}_i := c L_i(\mu_0^{(i)} + \sqrt{d_i} + \norm{\nabla \log \pi_{0}^{(i)}(0)}),
\end{align*}
where $\mu_0^{(i)} := (\E\norm{z_0^{(i)}}^2)^{1/2}$ and $c \geq 1$ is a universal constant.
Furthermore, we combine the individual constants together as 
$\mu_0 := \sqrt{\sum_{i=1}^{K} (\mu_0^{(i)})^{2}}$,
$\beta := \sqrt{\sum_{i=1}^{K} \beta_i^2}$, and 
$\mu_{t,x} := \mu_0 \vee \sigma_t \sqrt{D}$.
Finally, as before, we let $\mu_{x}^{(i)} := \lim_{t\to\infty}\mu_{t, x}^{(i)}$
and $\mu_{x} := \lim_{t\to\infty}\mu_{t, x}$.
Our first result mirrors that of \Cref{thm:score_estimation_latent}, and 
provides an error bound on the learned score functions
under the latent independent structure \eqref{eq:independent_components}.
\begin{restatable}{mythm}{scoreestimationindcomponents}
\label{thm:score_estimation_ind_components}
Suppose that $p_0$ follows the latent structure
\eqref{eq:independent_components}, and that both \Cref{assumption:sub_gaussian_z_0_components} and \Cref{assumption:lipschitz_score_functions} hold.
Fix a $t > 0$ and define 
\begin{align}
    \scrF_t := \{ s : \R^D \mapsto \R^D \mid \norm{s}_{\calF_1} \leq R_t \}, \quad R_t := \sum_{i=1}^{K} \bar{R}_{t}^{(i)} n^{\frac{d_i+1}{2(d_i+5)}}, \label{eq:F_t_ind_components}
\end{align}
where $\bar{R}_t^{(i)}$ does not depend on $n$.
Suppose that $n$ satisfies 
\begin{align}
    n \geq n_0(t) := \mathrm{poly}(D, 1/\sigma_t, \mu_{t,x} \vee \beta) \cdot \max_{i \in [K]} \mathrm{poly}_{d_i}(\bar{L}_i, \mu_{t,x}^{(i)} \vee \beta^{(i)}). \label{eq:ERM_score_matching_ind_components_burnin}
\end{align}
Then, the empirical risk minimizer $\hat{s}_t \in \argmin_{s \in \scrF_t} \hat{\calL}_t(s)$ satisfies:
\begin{align}
    \E_{\calD_t}[\calR_t(\hat{s}_t)] \leq \sum_{i=1}^{K} \tilde{O}_{d_i}(1) \left[  \frac{D^2 K}{\sigma_t^2 n}(\bar{L}_i (\mu_{t,x}^{(i)} \vee \beta_i))^{d_i+3} (\mu_{t,x} \vee \beta)^2 \right]^{\frac{2}{d_i+5}} + \tilde{O}(1) \sqrt{ \frac{D^2}{\sigma_t^4 n}(\mu_{t,x} \vee \beta)^2 }. \label{eq:ERM_score_matching_ind_components_bound}
\end{align}
\end{restatable}
Ignoring all parameters other than $K$ and $n$, we have that the risk scales as $\sum_{i=1}^{K} (K/n)^{2/(d_i+5)}$, 
which again captures the intrinsic dimensionality of the problem.
To the best of our knowledge, this is the first result establishing
a score function error bound in the setting \eqref{eq:independent_components}
which depends primarily on the latent $d_i$'s.
Our final result mirrors that of \Cref{corollary:sampling_subspace} and establishes an end-to-end sampling guarantee for this setting.
\begin{restatable}{mycorollary}{samplingindcomponents}
\label{corollary:sampling_ind_components}
Fix $\e, \zeta \in (0, 1)$.
Suppose that $p_0$ follows the latent structure
\eqref{eq:independent_components}, and that both \Cref{assumption:sub_gaussian_z_0_components} and \Cref{assumption:lipschitz_score_functions} hold.
Consider the exponential integrator \eqref{eq:exponential_integrator} with
$(N, T)$ as in \eqref{eq:diffusion_T_N_setting} and reverse process discretization timesteps $\{\tau_i\}_{i=0}^{N}$ defined as in \eqref{eq:reverse_process_timesteps}.
Next, define the forward process
timesteps $\{t_i\}_{i=0}^{N-1}$ by $t_i := T - \tau_{N-i}$.
Suppose the exponential integration scheme is run with score functions $\{ \hat{s}_{t_i} \}_{i=0}^{N-1}$,
where $\hat{s}_{t_i} \in \argmin_{s \in \scrF_{t_i}} \hat{\calL}_{t_i}(s)$ with
$\scrF_{t}$ as defined in \eqref{eq:F_t_ind_components}.
Suppose that $n \geq n_0(\zeta)$ satisfies:
\begin{align*}
    n \geq (\mu_x \vee \beta)^2 \max\left\{ \max_{i \in [K]} \left\{ \frac{\tilde{O}_{d_i}(1) D^2}{\zeta} K^{(d_i+7)/2} (\bar{L}_i (\mu_x^{(i)} \vee \beta))^{d_i+3} \cdot \e^{-(d_i+5)} \right\}, \frac{\tilde{O}(1) D^2}{\zeta^2}  \cdot \e^{-4}\right\} .
\end{align*}
where $n_0(\cdot)$ is defined in \eqref{eq:ERM_score_matching_ind_components_burnin}.
With constant probability (over the randomness of the training datasets $\{\calD_{t_i}\}_{i=0}^{N-1}$), we have that
$\KL{p_\zeta}{\mathrm{Law}(\hat{y}_{T-\zeta})} \leq \e^2$.    
\end{restatable}

We note that, unlike the latent subspace setting of \Cref{sec:results:subspace},
given \Cref{assumption:lipschitz_score_functions} it is possible to prove a bound on $\KL{p_0}{\mathrm{Law}(\hat{y}_{T-\zeta})}$ directly, since $\nabla \log p_t$ is uniformly Lipschitz for all $t \geq 0$. This can be done by using 
\citet[Theorem 2]{chen2023sampling} to analyze the backwards exponential integrator process \eqref{eq:exponential_integrator} instead of
the results of \citet{benton2024nearly}. We elect to utilize the latter's analysis in the interest of consistency with \Cref{sec:results:subspace} where it is required.

\subsubsection{Non-orthogonal independent components}
\Cref{sec:results:independent_components} shows that a diffusion model based on a shallow neural network can adapt to hidden independent component structure.
A natural question is whether this extends to the non-orthogonal case, similar to independent component analysis (ICA)~\citep{Herault_1985_ICA}. 
Here, we explain why a direct extension of our argument works at $t=0$ case, but breaks whenever $t > 0$ since the addition of noise breaks the independence structure.
We then use data whitening to address the issue.

Recall from \eqref{eq:independent_components} that $z_{0}=(z_{0}^{(1)},\dots,z_{0}^{(K)})$ where $x^{(i)}\in\mathbb{R}^{d_{i}}$
for $d_{1}+\dots+d_{K}=D$, and where each of the $K$ components are
sampled independently. 
Now, let us assume that $x_0 = A z_0$, where $A \in \R^{D \times D}$ is invertible, but not necessarily orthonormal.
Note that for $t=0$, the score function $\nabla \log p_0$ can be expressed as
$\nabla \log p_0(x) = \sum_{i=1}^{K} A^{-\T} P_i^\T \nabla \log \pi_0^{(i)}(P_i A^{-1} x)$,
where $P_i  \in \R^{d_i \times D}$ selects the $d_i$ coordinates that correspond to the $i$-th variable group.
Hence, the score function has the structure of a sum of low-index functions, and the $\calF_{1}$ norm of the score can be bounded in terms of the sum of $\calF_{1}$ norms of each component.
However, once isotropic noise is added, this structure is lost in general because the isotropy is not preserved in the latent space.
That is, $z_{t}=A^{-1}(m_{t}x_{0}+\sigma_{t}w)=m_{t}z_{0}+\sigma_{t}A^{-1}w$.
In this case, the sum of low-index
functions structure is not preserved.%

A possible fix for this issue is to first whiten the data, and then to apply the results of \Cref{sec:results:independent_components} to the whitened data.
Specifically, write $\mathrm{Cov}(x_0) = A \Sigma A^\T$, where $\Sigma$ is a block diagonal matrix with $K$ blocks $\Sigma^{(1)},\dots,\Sigma^{(K)}$.
Now consider the transform $\bar{x}_0 := \Sigma^{-\frac{1}{2}} A^{-1} x_0$, which
orthogonalizes the components and whitens
each of them independently. Indeed, $\bar{x}_0$ follows the latent structure \eqref{eq:independent_components}, and hence the results from \Cref{sec:results:independent_components} directly apply to $\bar{x}_0$.
However, one caveat with this approach is that the whitening factor $\Sigma^{-1/2} A^{-1}$ must be learned from the data samples $x_0^i \sim p_0(\cdot)$.
Standard results in covariance estimation~\citep[see e.g.,][Chapter 6]{wainwright2019high} allow us to learn $\Sigma^{-1/2} A^{-1}$ (modulo rotation) up to $n^{-1/2}$ accuracy. 
It is an interesting question that we leave to future work to study how the associated estimation error propagates through both the learning and sampling procedures when obtaining a final sample complexity bound.

\section{Proof Ideas}
\label{sec:proof_ideas}
Here, we outline the key proof ideas behind the results in
\Cref{sec:results:subspace} and
\Cref{sec:results:independent_components}.
We focus our discussion exclusively on bounding the
error of the score function estimate, as translating score error 
into sample quality bounds is already well-established in the literature (cf.~\Cref{sec:related_work}).
For this discussion, we fix a specific value of $t > 0$, noting from our discussion in \Cref{sec:results} (specifically~\Cref{eq:dsm_empirical_loss_single_scale}) that we learn separate score models for a fixed sequence of 
forward process timesteps $\{t_i\}_{i=0}^{N-1}$.

\subsection{Basic inequality}
Recall that $s_t \in \argmin_{s \in \scrF_t} \hat{\calL}_t(s)$
is the empirical risk minimizer (ERM) of the empirical denoising loss 
$\hat{\calL}_t$ over the function class $\scrF_t = \{ s : \R^D \mapsto \R^D \mid \norm{s}_{\calF_1} \leq R_t \}$, where the norm bound $R_t$ will be determined.
Our first step uses the link between the 
$L_2(p_t)$ score error $\calR_t(s)$ and the DSM loss $\calL_t(s)$
in addition to standard arguments from the analysis of ERM to show 
the following basic inequality
for all $\e \geq 0$:
\begin{align}
    \E_{\calD_t}[\calR_t(\hat{s}_t)] \leq (1+\e) \inf_{s \in \scrF_t} \calR_t(s) + \E_{\calD_t} \sup_{s \in \scrF_t}[ \calL_t(s) - (1+\e) \hat{\calL}_t(s) ] + \e \cdot C_t, \label{eq:DSM_basic_inequality} 
\end{align}
where $C_t := \E \Tr\mathrm{Cov}( \sigma_t^{-2}(m_t x_0 - x_t) \mid x_t )$.
The basic inequality \eqref{eq:DSM_basic_inequality}
contains three key terms: 
an \emph{approximation-theoretic} term $T_1 := (1+\e) \inf_{s \in \scrF_t} \calR_t(s)$ which measures how well the $\calF_1$-norm bounded subset $\scrF_t$ approximates the true score,
a \emph{uniform convergence} term $T_2 := \E_{\calD_t} \sup_{s \in \scrF_t}[ \calL_t(s) - (1+\e) \hat{\calL}_t(s) ]$ 
over the function class $\scrF_t$, and
a third \emph{offset} term $T_3 := \e \cdot C_t$ which
trades off a fast rate for $T_2$ (controlled via $\e$) with
the constant offset $C_t$ between the score error and the DSM loss. 
In addition, the
first two terms are in tension with each other, and must be carefully balanced to achieve the desired rate.

\subsection{Approximation of structured models with $\calF_1$}
\label{sec:proof_ideas:F1_approx}

Our analysis is based on careful control of the approximation error term $T_1$ in \eqref{eq:DSM_basic_inequality} of the structured models we consider in a way such that the requisite $\calF_1$-norm $R_t$ does not depend exponentially on the ambient dimension $D$.
This is accomplished by first understanding the low-dimensional structure present in the score functions $\nabla \log p_t$, and then arguing that this low-dimensional structure
can be approximated with a norm bound $R_t$ that depends reasonably on $D$.

\paragraph{Subspace structure.}
We first consider the low dimensional subspace from \Cref{sec:results:subspace}.
Under this model, we have the following expression relating the score $\nabla \log p_t$ to the score $\nabla \log \pi_t$ of the latent $z_t$.
\begin{myprop}[{see e.g.,~\citet[Lemma 1]{chen2023subspace}}]
\label{prop:subspace_structure_score_function}
The following decomposition holds under the subspace model \eqref{eq:subspace_structure}:
\begin{align}
    \nabla \log p_t(x) = U \nabla \log \pi_t( U^\T x ) - \frac{1}{\sigma_t^2} (I-UU^\T) x. \label{eq:subspace_structure_score_function}
\end{align}
\end{myprop}

\paragraph{Independent components.}
We next consider the independent structure
from \Cref{sec:results:independent_components}.
Under this model, we have the following decomposition for the score.
\begin{myprop}
\label{prop:independent_components_score_function}
The following holds under the independent components model \eqref{eq:independent_components}:
\begin{align}
    \nabla \log p_t(x) = \sum_{i=1}^{K} U P_i^\T \nabla \log \pi_t^{(i)}( P_i U^\T x ), \label{eq:independent_components_score_function}
\end{align}
where $P_i \in \R^{d_i \times D}$
selects the coordinates corresponding to the $i$-th variable group.
\end{myprop}
Note that the proof of 
\Cref{prop:independent_components_score_function} follows directly from the standard 
change of variables formula, and the fact that the distribution of $w$ is unchanged when pre-multiplied by $U^\T$.

The score function decompositions
\eqref{eq:subspace_structure_score_function}
and \eqref{eq:independent_components_score_function} both exhibit similar structure, where 
latent score functions are embedded into 
a score function in the ambient space via a linear encoding/decoding process.
Fortunately, this embedding preserves the $\calF_1$-norm of the underlying function.
\begin{myfact}
\label{fact:F1_norm_equiv}
Let $f : \R^d \mapsto \R^d$ have bounded $\calF_1$-norm, and let $U \in O(D, k)$.
Consider $g : \R^D \mapsto \R^D$ defined as $g(x) = U f(U^\T x)$. We have that
$\norm{g}_{\calF_1} = \norm{f}_{\calF_1}$.
\end{myfact}
\noindent
Hence, if the latent function $f : \R^d \to \R^d$ can be approximated well in $\calF_1$, then the embedded function is also approximated well with the \emph{same} $\calF_1$ norm. 
This is the key observation that enables our results.
Concretely, suppose that
$\hat{f} : \R^d \mapsto \R^d$ has bounded $\calF_1$ norm and approximates
$f : \R^d \to \R^d$ via
$\sup_{z \in B_2(d, M)} \norm{f(z) - \hat{f}(z)} \leq \e$.
Then, $\hat{g}(x) = U \hat{f}(U^\T x)$ approximates $g(x) = U f(U^\T x)$ via
$\sup_{x \in B_2(D, M)} \norm{g(x) - \hat{g}(x)} \leq \e$, and $\norm{\hat{g}}_{\calF_1} = \norm{\hat{f}}_{\calF_1}$.

It remains to argue that $\calF_1$ can approximate low dimensional functions well. 
Fortunately, the approximation properties of $\calF_1$ functions over various function classes is well-understood~\citep{bach2017breaking,jacot2024dnns}.
In particular, we utilize the following $\calF_1$ approximation result for Lipschitz continuous functions, adopted from \citet[Proposition 6]{bach2017breaking}.
\begin{restatable}{mylemma}{lipschitzapprox}
\label{lemma:F1_lipschitz_approx}
Let $f : \R^d \to \R^d$ be $L$-Lipschitz and $B$-bounded on $B_2(d, M)$. 
Define $K := B \vee LM$.
For any $\e \in (0, K/2)$, there exists an $f_\e \in \calF_1$ such that
$\sup_{x \in B_2(d, M)} \norm{f(x) - f_\e(x)} \leq \e$ and:
\begin{align*}
    \norm{f_\e}_{\calF_1} \leq O_d(1) K \left(\frac{K}{\e}\right)^{(d+1)/2} \log^{(d+1)/2} \left( \frac{K}{\e} \right).
\end{align*}
\end{restatable}

\subsection{Uniform convergence of the DSM loss}

With the approximation result in place, we turn to the analysis of uniform convergence term $T_2$ in \eqref{eq:DSM_basic_inequality}.
Since the DSM loss is a least-squares regression problem, we can utilize existing results for analyzing generalization error with smooth losses~\citep{srebro2010fastrates}.
However, the main technical hurdle here is dealing with the fact that the data tuples $(x_0, x_t)$ are not uniformly bounded, which is a technical assumption needed in many of these arguments.\footnote{As an alternative to boundedness, one could also rely on small-ball arguments~\citep{mendelson2015learning}. In the interest of keeping our assumptions minimal as possible, we do not pursue this approach.} 
While this can be handled straightforwardly for a fixed time $t$ via standard truncation arguments, one challenge is ensuring that the resulting high probability bounds degrade nicely as $t \to 0$.
The reason this is necessary is because the smallest timescale $t_0$ used will ultimately scale with the number of datapoints $n$.

To highlight the class of issues that arise in our truncation arguments, 
consider the latent score function $\nabla \log \pi_t$ arising from the subspace structure setting \eqref{eq:subspace_structure}. 
In order to apply a truncation argument for analyzing $T_2$, 
we need to argue that $\norm{\nabla \log \pi_t(z)}$ is bounded uniformly over a high-probability truncation set $z \in B_2(d, M)$.
By the continuity of $\nabla \log \pi_t$, we know that $\sup_{z \in B_2(d, M)} \norm{\nabla \log \pi_t(z)} = A_t < \infty$.
However, we need to control the behavior of $A_t$ as $t \to 0$.
By leveraging the perturbation analysis of \cite{lee2023convergence}, we show that for all $t \geq 0$, under our assumptions, the inequality
$\norm{\nabla \log \pi_t(z)} \leq \bar{L}(1 + \norm{z})$ holds for all $z \in \R^d$.
Hence, we can bound $A_t \leq \bar{L} (1+M)$ for all $t \geq 0$.

\section{Conclusion}
\label{sec:conclusion}

In this work, we showed that diffusion models based on shallow neural networks 
applied to data from distributions that contain low dimensional structure -- specifically, linear subspace and hidden independent component structure -- exhibit favorable sample complexity bounds that primarily depend on the
latent dimensionality of the problem, thereby avoiding the curse of dimensionality.
We accomplish this by leveraging the low-index structure of the Barron space,
which allows us to avoid specific latent-aware architectural modifications
and computationally intractable sparsity constraints, both of which have been used to obtain similar results in earlier work.

Several exciting future research threads arise directly from our study.
The most pertinent direction is to increase the scope of the latent structures covered by our analysis, to include, for example, non-linear manifolds.
Another related question is whether or not favorable results that avoid 
the curse of dimensionality can be shown for latent diffusion models~\citep{rombach2022latentdiffusion}, which first learn an autoencoder
before learning a diffusion model in the autoencoder's latent space.
On the algorithmic front, an interesting open question is whether or not gradient-based optimization algorithms can efficiently learn the low-index structure associated with the latent models studied in this paper.
Finally, improving our rates to match the minimax optimal score estimation rates of~\citet{wibisono2024optimal} -- with the latent dimension playing the role of the ambient dimension -- is another exciting area for future work.

\subsubsection*{Acknowledgments}
Ingvar Ziemann acknowledges support by a Swedish Research Council international postdoc grant.
\bibliographystyle{unsrtnat}
\bibliography{paper}

\newpage
\appendix

\section{Preliminary results}
\label{sec:appendix:preliminary}

We first make explicit the relation between the
score matching error $\calR_t$ and the denoising loss $\calL_t$.
In the sequel, we will make frequent use of 
Tweedie's formula~\citep[see e.g.][]{efron2011tweedie}:
\begin{align*}
    \nabla \log p_t(x_t) = \E\left[\nabla \log q_t(x_t \mid x_0) \mid x_t \right], \quad \nabla \log q_t(x_t \mid x_0) := \frac{m_t x_0-x_t}{\sigma_t^2}.
\end{align*}

\begin{myfact}[DSM loss minimizes score]
\label{fact:dsm_score_matching_loss}
We have that for all sufficiently regular $s : \R^D \mapsto \R^D$:
\begin{align}
    \calR_t(s) = \calL_t(s) - C_t,
\end{align}
where the offset constant $C_t$ is given by:
$$
    C_t := \E \Tr\mathrm{Cov}\left( \frac{m_t x_0 - x_t}{\sigma_t^2} \mid x_t \right).
$$
Furthermore, we can bound $C_t$ by:
$$
    C_t \leq D/\sigma_t^2.
$$
\end{myfact}
\begin{proof}
The standard proof for denoising score matching~\citep[see e.g.][Section 4.2]{vincent2011denoising}
shows that for all $s$:
$$
    \calR_t(f) = \calL_t(f) + \E_{(x_0, x_t)}[ \norm{\nabla \log p_t(x_t)}^2 - \norm{\nabla \log q_t(x_t \mid x_0)}^2 ].
$$
Hence,
\begin{align*}
    &\E_{(x_0, x_t)}[ \norm{\nabla \log p_t(x_t)}^2 - \norm{\nabla \log q_t(x_t \mid x_0)}^2 ] \\
    &= \E_{(x_0, x_t)}\left[ \bignorm{\E\left[\frac{m_t x_0-x_t}{\sigma_t^2} \mid x_t \right] }^2 - \bignorm{ \frac{m_t x_0-x_t}{\sigma_t^2} }^2  \right] \\
    &= - \E \E\left[  \bignorm{ \frac{m_t x_0-x_t}{\sigma_t^2} }^2 - \bignorm{\E\left[\frac{m_t x_0-x_t}{\sigma_t^2} \mid x_t \right] }^2 \mid x_t \right] \\
    &= - \E \Tr\mathrm{Cov}\left( \frac{m_t x_0-x_t}{\sigma_t^2} \mid x_t \right) \\
    &= - C_t.
\end{align*}
To bound $C_t$, we observe:
\begin{align*}
    C_t = \E \Tr\mathrm{Cov}\left( \frac{m_t x_0-x_t}{\sigma_t^2} \mid x_t \right) &= \E \Tr\mathrm{Cov}\left( \frac{w}{\sigma_t} \mid x_t \right) \leq \E \norm{w/\sigma_t}^2 = D/\sigma_t^2.
\end{align*}
\end{proof}

We next state a result from \cite{benton2024nearly}
regarding the quality of the samples generated via the exponential integrator
scheme \eqref{eq:exponential_integrator}.
\begin{mylemma}[{Sampler quality from $L_2$ score bounds~\citep[Theorem 2]{benton2024nearly}}]
\label{thm:sampler_quality_from_L2_score}
Fix a $T \geq 1$ and $\zeta \in (0, 1)$.
Also fix an $N \in \N_+$ which is even and satisfies $N \geq 2 \log(1/\zeta)$.
Define a sequence of strictly increasing backwards process times $\{\tau_i\}_{i=0}^{N}$:
\begin{align}
    \tau_i = \begin{cases} 
        2(T-1)\frac{i}{N} &\text{if } i \in \{0, \dots, N/2\}, \\
        T - \zeta^{2i/N-1} &\text{if } i \in \{N/2+1,\dots,N\}.
        \end{cases}
\end{align}
Let $\gamma_i := \tau_{i+1}-\tau_i$ for $i \in \{0, \dots, N-1\}$.
Suppose we have $N$ score functions $\hat{s}_t(x)$
for $t \in \{T - \tau_i\}_{i=0}^{N-1}$
which satisfy:
\begin{align}
    \sum_{i=0}^{N-1} \gamma_i \E_{p_{T-\tau_i}} \norm{ \hat{s}_{T-\tau_i} - \nabla \log p_{T-\tau_i}}^2 \leq \e^2_{\mathrm{score}}. \label{eq:weighted_L2_score_requirement}
\end{align}
Then, we have the following guarantee
for the exponential integrator \eqref{eq:exponential_integrator}:
\begin{align*}
    \KL{p_{\zeta}}{\mathrm{Law}(\hat{y}_{T-\zeta})} \lesssim \e^2_{\mathrm{score}} + \kappa^2 D N + \kappa (DT + \mu_0^2) + (D+\mu_0^2) e^{-2T},
\end{align*}
where $\kappa, \mu_0$ are defined as:
\begin{align*}
    \kappa := \frac{2(T-1)+4\log(1/\zeta)}{N}, \quad \mu_0 := \E \norm{x_0}^2.
\end{align*}
\end{mylemma}
\begin{proof}
In order to apply \citet[Theorem 2]{benton2024nearly}, 
we need to compute a $\kappa$ such that:
$$
    \gamma_k \leq \kappa \min\{1, T-t_{k+1}\}, \quad \forall k \in \{0, \dots, N-1\}.
$$
To do this, we follow the proof of
\citet[Corollary 1]{benton2024nearly}.
First, for $k \in \{0, \dots, N/2-1\}$, we have that
$t_{k+1} \leq T-1$, and hence $T-t_{k+1} \geq T - (T-1) = 1$.
Hence, we can simply take $\kappa \geq 2(T-1)/N$.
Now, for $k \in \{N/2, \dots, N-1\}$,
we have that $t_{k+1} \geq T-1$, and therefore
$T-t_{k+1} \leq T - (T-1) = 1$.
Hence we need a $\kappa$ such that
$\gamma_k \leq \kappa (T-t_{k+1})$.
We therefore compute:
\begin{align*}
    \zeta^{2k/N-1} - \zeta^{2(k+1)/N-1} = \gamma_k \leq \kappa (T-t_{k+1}) = \kappa \zeta^{2(k+1)/N-1}.
\end{align*}
From this we see that $\kappa \geq \zeta^{-2/N} - 1$ is required.
A sufficient condition takes:
\begin{align*}
    \zeta^{-2/N} - 1 &= \exp\left(\frac{2}{N} \log(1/\zeta) \right) - 1 \\
    &\leq 1 + (e-1) \frac{2}{N} \log(1/\zeta) - 1 &&\text{since } e^x \leq 1 + (e-1) x \text{ for } x \in [0, 1] \\
    &\leq \frac{4\log(1/\zeta)}{N}.
\end{align*}
Hence, in total we can set:
\begin{align*}
    \kappa = \frac{2(T-1) + 4\log(1/\zeta)}{N}.
\end{align*}
The result now follows from invoking \citet[Theorem 2]{benton2024nearly}.
\end{proof}
Note that an immediate consequence of \Cref{thm:sampler_quality_from_L2_score} is the following observation: for any $\e \in (0, 1)$, setting
\begin{align*}
    T = c_0 \log\left(\frac{\sqrt{D} \vee \mu_0}{\e}\right), \quad N = 2 \left\lceil c_1 \frac{D \vee \mu_0^2}{\e^2} \left[  \log^2\left(\frac{\sqrt{D} \vee \mu_0}{\e}\right) + \log^2\left(\frac{1}{\zeta}\right) \right] \right\rceil, %
\end{align*}
for some universal positive constants $c_0, c_1$,
we have that
\begin{align*}
    \KL{p_\zeta}{ \mathrm{Law}(\hat{y}_{T-\zeta}) } \lesssim \e^2_{\mathrm{score}} + \e^2.
\end{align*}

Next, we state a technical lemma regarding a perturbation result for
Gaussian mollifications at small scale.
\begin{mylemma}[{Score function perturbation~\citep[Lemma C.12]{lee2023convergence}}]
\label{prop:mollification_comparison}
Suppose that $p(x)$ is a density on $\R^d$ such that $\nabla \log p(x)$ is $L$-Lipschitz on $\R^d$. 
For $\alpha \geq 1$, define
the corresponding density $p_\alpha(x) := \alpha^d p(\alpha x)$. Let $\gamma_{\sigma^2}$ denote an ${\mathsf N}(0, \sigma^2 I_d)$ distribution. If $L \leq 1/(2\alpha^2 \sigma^2)$, then we have the following bound for all $x \in \R^d$:
\begin{align*}
    &\norm{ \nabla \log p(x) - \nabla \log p_\alpha \ast \gamma_{\sigma^2}(x) } \\
    &\qquad\leq 6 \alpha^2 L \sigma \sqrt{d} + (\alpha + 2\alpha^3 L \sigma^2)(\alpha - 1) L \norm{x} + (\alpha-1 + 2\alpha^3 L \sigma^2) \norm{\nabla \log p(x)}.
\end{align*}
\end{mylemma}

The previous mollification lemma is next used to bound the 
score functions uniformly over time.
\begin{myprop}
\label{prop:linear_Linf_bound}
Consider a forward diffusion process $z_t \overset{\mathsf d}{=} m_t z_0 + \sigma_t w$ on $\R^d$. Let $\pi_t$ denote the marginal distribution of $z_t$ for all $t \geq 0$. Suppose that $\nabla \log \pi_t$ is $L$-Lipschitz (for $L \geq 1$) on $\R^d$ for all $t \geq 0$.
Defining $\bar{L} := c L(\E\norm{z_0} + \sqrt{d} + \norm{\nabla \log \pi_0(0)})$ where $c \geq 1$ is an universal constant, 
we have that for all $t \geq 0$ and $z \in \R^d$,
\begin{align*}
    \norm{\nabla \log \pi_t(z)} \leq \bar{L}(1 + \norm{z}).
\end{align*}
\end{myprop}
\begin{proof}
Define $\bar{\pi}_t(z) := m_t^{-d} \pi_0(m_t^{-1} z)$,
which is the density of the random variable $m_t z_0$.
Hence, we have that $\pi_t = \bar{\pi}_t \ast \gamma_{\sigma_t^2}$.
By \Cref{prop:mollification_comparison},
whenever $L \leq m_t^2/(2\sigma_t^2)$, we have that:
\begin{align*}
    \norm{ \nabla \log \pi_t(0) } \leq 6 m_t^{-2} L \sigma_t \sqrt{d} + (m_t^{-1} + 2 m_t^{-3} L \sigma_t^2) \norm{\nabla \log \pi_0(0)}.
\end{align*}
We now compute the range of $t$'s for which
$L \leq m_t^2/(2\sigma_t^2)$ holds.
Using the specific form of $m_t, \sigma_t$,
\begin{align*}
    2L \leq \frac{m_t^2}{\sigma_t^2} = \frac{\exp(-2t)}{1-\exp(-2t)} = \frac{1}{\exp(2t) - 1} \Longleftrightarrow t \leq \frac{1}{2} \log\left(1+\frac{1}{2L}\right) =: t_\star.
\end{align*}
First, note that since we assume $L \geq 1$,
then we have that $t_\star$ is bounded by a universal constant.
Hence, for $t \leq t_\star$,
\begin{align*}
    \norm{\nabla \log \pi_t(0)} \lesssim L (\sqrt{d} + \norm{\nabla \log \pi_0(0)}).
\end{align*}
Hence for any $z$,
\begin{align*}
    \norm{\nabla \log \pi_t(z)} &\leq L\norm{z} + \norm{\nabla \log \pi_t(0)} \\
    &\leq L\norm{z} + c L(\sqrt{d} + \norm{\nabla \log \pi_0(0)}) \\
    &\leq cL(\sqrt{d} + \norm{\nabla \log \pi_0(0)})(1 + \norm{z}).
\end{align*}
On the other hand, when $t \geq t_\star$,
more work is needed.
When $t \geq t_\star$, we first bound:
\begin{align*}
    \sigma_t^{-1} \leq \sigma_{t_\star}^{-1} = \sqrt{2L+1}.
\end{align*}
Define two events:
\begin{align*}
    \calE_1 := \left\{ \E[ \norm{w}/\sigma_t \mid z_t] \geq 4 \sqrt{d}/\sigma_t \right\}, \quad \calE_2 := \left\{ \norm{z_t} \geq 4(\E\norm{z_0} + \sigma_t\sqrt{d})  \right\}
\end{align*}
By Markov's inequality we have that:
\begin{align*}
    \Pr_{z_t}\left\{ \calE_1 \right\} \leq 1/4, \quad \Pr_{z_t}\left\{ \calE_2 \right\} \leq 1/4.
\end{align*}
Now suppose that $\calE_1^c \subseteq \calE_2$. Then, we have a contradiction, since:
\begin{align*}
    1 - 1/4 \leq \Pr_{z_t}\{ \calE_1^c \} \leq \Pr_{z_t}\{ \calE_2 \} \leq 1/4.
\end{align*}
Hence, there must exists an $\omega \in \calE_1^c$ such that $\omega \not \in \calE_2$. 
Hence, there exists a $\bar{z}_t$ satisfying:
\begin{align*}
    \norm{\nabla \log \pi_t(\bar{z}_t)} \leq \E[ \norm{w}/\sigma_t \mid z_t=\bar{z}_t] \leq 4 \sqrt{d}/\sigma_t , \quad \norm{\bar{z}_t} \leq 4(\E\norm{z_0} + \sigma_t\sqrt{d}).
\end{align*}
Hence for any $z$, 
\begin{align*}
    \norm{\nabla \log \pi_t(z)} &\leq L \norm{z - \bar{z}_t} + \norm{\nabla \log \pi_t(\bar{z}_t)} \\
    &\leq L \norm{z} + 4 L (\E\norm{z_0} + \sigma_t\sqrt{d}) + 4\sqrt{d}/\sigma_t \\
    &\leq L \norm{z} + 4 L (\E\norm{z_0} + \sqrt{d}) + 4\sqrt{(2L+1)d} \\
    &\leq c' L( \E\norm{z_0} + \sqrt{d} )(1 + \norm{z}).
\end{align*}
\end{proof}

Next, we state a $L_\infty$ approximation result for scalar-valued Lipschitz function from \citet{bach2017breaking}.
\begin{mylemma}[{$L_\infty$ approximation of scalar Lipschitz functions~\citep[Proposition 6]{bach2017breaking}}]
\label{lemma:bach_lipschitz_approx}
Let $f : \R^d \mapsto \R$.
Suppose that $f$ is $L$-Lipschitz and $B$-bounded on $B_2(d, M)$.
Set $K := B \vee LM$.
For any $\gamma \geq O_d(1) \cdot K$ there exists an $f_\gamma \in \calF_1$ such that $\norm{f_\gamma}_{\calF_1} \leq \gamma$ and:
\begin{align*}
    \sup_{x \in B_2(d, M)} \abs{f(x)-f_\gamma(x)} \leq O_d(1) K \left(\frac{K}{\gamma}\right)^{2/(d+1)} \log\left(\frac{\gamma}{K}\right).
\end{align*}
\end{mylemma}

Our next result is a simple technical fact which we will utilize in our
truncation analysis.
\begin{myprop}
\label{prop:random_variable_S_Scheck}
Let $S, \check{S}$ be two $\sfX$-valued random
variables over the same probability space.
Let $f : \sfX \mapsto \R$ be a measurable function.
We have:
\begin{align*}
    \E[ f(S) ] \leq \E[ f(\check{S}) ] + (\sqrt{ \E[f^2(S)]} + \sqrt{\E[f^2(\check{S})]})\sqrt{\Pr\{ S \neq \check{S}\}}.
\end{align*}
Note if $f$ is non-negative, then we have the simpler bound:
\begin{align*}
    \E[f(S)] \leq \E[f(\check{S})] + \sqrt{\E[ f^2(S) ] \Pr\{ S \neq \check{S} \}}.
\end{align*}
\end{myprop}
\begin{proof}
We have:
\begin{align*}
    \E[f(S)] &= \E[f(S) \ind\{ S = \check{S} \} ] + \E[ f(S) \ind\{S \neq \check{S}\} ] \\
    &= \E[f(\check{S}) \ind\{S = \check{S}\} ] + \E[ f(S) \ind\{S \neq \check{S}\} ] \\
    &= \E[f(\check{S})] + \E[ f(S) \ind\{S \neq \check{S}\} ] - \E[ f(\check{S}) \ind\{ S \neq \check{S} \} ] \\
    &\leq \E[ f(\check{S}) ] + \sqrt{\E[ f^2(S) ] \Pr\{ S \neq \check{S} \} } + \sqrt{\E[ f^2(\check{S}) ] \Pr\{ S \neq \check{S} \} }.
\end{align*}
\end{proof}

The next result is a simple algebraic fact which will be useful for solving for implicit inequalities involving logarithms.
\begin{myprop}[{Log dominance rule,~\citep[see e.g.][Lemma F.2]{du2021bilinearclasses}}]
\label{prop:log_dominance_rule}
Let $a, b, \nu$ be positive scalars.
Put $\bar{\nu} := (1+\nu)^{\nu}$.
Then,
\begin{align*}
    m \geq \bar{\nu} a \log^{\nu}(\bar{\nu} ab) \Longrightarrow m \geq a \log^{\nu}(b m).
\end{align*}
\end{myprop}

Next, we have an intermediate result to bound the Rademacher
complexity of $\calF_1$-norm bounded functions.
\begin{myprop}
\label{lemma:relu_vector_rademacher}
Let $\norm{x_i} \leq 1$ for $i \in [n]$. We have:
\begin{align*}
    \E_{\{\e_i\}} \sup_{u, v \in B_2(D)} \left|\frac{1}{n}\sum_{i=1}^n  \ip{u}{\e_i} \sigma(\ip{v}{x}) \right| \leq c \sqrt{\frac{D}{n}},
\end{align*}
where the $\e_i \in \{\pm 1\}^d$ are independent Rademacher random vectors\footnote{That is, each coordinate of $\e_i \in \{\pm 1\}^d$ is an independent Rademacher random variable.}
and $c > 0$ is a universal constant.
\end{myprop}
\begin{proof}
Define $X_{u,v} := \frac{1}{n}\sum_{i=1}^n \ip{u}{\e_i} \sigma(\ip{v}{x})$.
Observe that for $u_i, v_i \in B_2(D)$ for $i \in \{1,2\}$,
\begin{align*}
    X_{u_1, v_1} - X_{u_2, v_2} &= \frac{1}{n}\sum_{i=1}^{n} [ \sigma(\ip{v_1}{x_i}) - \sigma(\ip{v_2}{x_i}) ] \ip{u_1}{\e_i} + \frac{1}{n}\sum_{i=1}^{n} \sigma(\ip{v_2}{x_i}) \ip{u_1-u_2}{\e_i} \\
    &=: T_1 + T_2.
\end{align*}
First, we recall that a Rademacher random variable is $1$-sub-Gaussian, and therefore $\ip{u_i}{\e_i}$ is also $1$-sub-Gaussian since $\norm{u_i} \leq 1$.
Using the fact that ReLU is $1$-Lipschitz followed by Cauchy-Schwarz
and the assumption that $\norm{x_i} \leq 1$,
\begin{align*}
    \abs{ \sigma(\ip{v_1}{x_i}) - \sigma(\ip{v_2}{x_i}) } \leq \abs{ \ip{v_1-v_2}{x_i} } \leq \norm{v_1-v_2}.
\end{align*}
Hence, $[ \sigma(\ip{v_1}{x_i}) - \sigma(\ip{v_2}{x_i}) ] \ip{u_1}{\e_i}$ is $\norm{v_1-v_2}$-sub-Gaussian.
Consequently, $T_1$ is $\norm{v_1-v_2}/\sqrt{n}$-sub-Gaussian.
Similarly, since $\abs{ \sigma(\ip{v_2}{x_i}) } \leq 1$, we also have that
$T_2$ is $\norm{u_1-u_2}/\sqrt{n}$-sub-Gaussian.
Hence, the sum $T_1 + T_2$ is sub-Gaussian with constant:
$$
    \sqrt{2(\norm{v_1-v_2}^2 + \norm{u_1-u_2}^2)}/\sqrt{n}.
$$
Letting $\omega = (u, v)$, we consider the following metric on
$\Omega := B_2(D) \times B_2(D)$:
$$
    d((u_1, v_1), (u_2, v_2)) = \sqrt{ \norm{u_1-u_2}^2 + \norm{v_1-v_2}^2 }.
$$
Hence for any $\omega_1, \omega_2 \in \Omega$, 
the difference $X_{\omega_1} - X_{\omega_2}$ is $\sqrt{2/n} \cdot d(\omega_1, \omega_2)$-sub-Gaussian.
Therefore we can use Dudley's inequality~\citep[see e.g.][Chapter 8]{vershynin2018high} to bound:
\begin{align*}
    \E \sup_{\omega \in \Omega} X_\omega &\leq c n^{-1/2} \int_0^\infty \sqrt{ \log N( \e; \Omega, d  ) }\,\rmd \e 
    =  c n^{-1/2} \int_0^{\sqrt{2}} \sqrt{ \log N( \e; \Omega, d  ) }\,\rmd \e.
\end{align*}
Next, fix an $\e > 0$ and let $\{ u_i \}, \{ v_i \}$ be $\e/\sqrt{2}$-covers of $B_2(D)$.
Let $[u]$ (resp.~$[v]$) denote the closest point in the cover to $u$ (resp.~$v$).
Given $(u, v) \in \Omega$, we have:
\begin{align*}
    d((u, v), ([u], [v])) = \sqrt{ \norm{u - [u]}^2 + \norm{v - [v]^2 }} \leq \e.
\end{align*}
Using the standard volume estimate of the covering of $B_2(D)$~\citep[see e.g.][Chapter 4]{vershynin2018high},
\begin{align*}
    \log N(\e; \Omega, d) \leq 2D \log(1+2\sqrt{2}/\e).
\end{align*}
Consequently,
\begin{align*}
    \E \sup_{\omega \in \Omega} X_w \leq cn^{-1/2} \sqrt{2D} \int_0^{\sqrt{2}} \sqrt{\log(1+2\sqrt{2}/\e)}\,\rmd\e = c' \sqrt{D/n}.
\end{align*}
\end{proof}

Our final preliminary result translates the previous bound
\Cref{lemma:relu_vector_rademacher} to a bound on
the Rademacher complexity of $\calF_1$-norm balls.
\begin{myprop}
\label{lemma:rademacher_complexity_F1}
Let $\scrF = \{ s : \R^d \mapsto \R^d \mid \norm{s}_{\calF_1} \leq R \}$. 
For any $\check{x}_i \in B_2(D, M)$, $i \in [n]$, we have:
\begin{align*}
    \E_{\{\e_i\}} \sup_{f \in \scrF} n^{-1} \bigabs{\sum_{i=1}^{n} \ip{\e_i}{f(\check{x}_i)}} \leq c R M \sqrt{\frac{D}{n}},
\end{align*}
where the $\e_i \in \{\pm 1\}^d$ are independent Rademacher random vectors 
and $c > 0$ is a universal constant.
\end{myprop}
\begin{proof}
For any $\{\e_i\}$ and $f \in \scrF$, observe that by the definition of $\calF_1$:
\begin{align*}
    n^{-1} \bigabs{\sum_{i=1}^{n} \ip{\e_i}{f(\check{x}_i)}} &= n^{-1} \bigabs{\sum_{i=1}^{n} \bigip{\e_i}{\int u \sigma(\ip{v}{\check{x}_i}) \,\rmd \mu(u, v)}} \\
    &= \bigabs{\int \left[ \frac{1}{n}\sum_{i=1}^{n} \ip{\e_i}{u} \sigma(\ip{v}{\check{x}_i}) \right] \rmd\mu(u, v)} \\
    &\leq R \sup_{u,v \in \mathbb{S}^{D-1}} \bigabs{\frac{1}{n}\sum_{i=1}^{n} \ip{\e_i}{u} \sigma(\ip{v}{\check{x}_i})  }.
\end{align*}
Hence,
\begin{align*}
    \E_{\{\e_i\}} \sup_{f \in \scrF} n^{-1} \bigabs{\sum_{i=1}^{n} \ip{\e_i}{f(\check{x}_i)}} \leq R \cdot \E_{\{\e_i\}}\sup_{u,v \in \mathbb{S}^{D-1}} \bigabs{\frac{1}{n}\sum_{i=1}^{n} \ip{\e_i}{u} \sigma(\ip{v}{\check{x}_i}) },
\end{align*}
from which the claim follows by 
\Cref{lemma:relu_vector_rademacher}
and using homogeneity of ReLU to scale the data points $\check{x}_i$ to $B_2(D)$.
\end{proof}

\section{$\calF_1$ approximation theory for Lipschitz continuous functions}
\label{sec:appendix:F1_approximation}

Here we develop the necessary results to establish that $\calF_1$ functions can approximate structured score functions in an efficient way. 
Our first result is a preliminary result that allows us to translate $L_\infty$ approximation bounds to $L_2$ bounds.
For what follows, let the notation $\calP(\sfX)$ denote the set of subsets of $\sfX$.
\begin{myprop}
\label{prop:score_function_approx_set_version}
Let $M : (0, 1) \mapsto \calP(\R^D)$ be such that:
$$
    \forall\,\delta \in (0, 1), \quad \Pr_{x_t \sim p_t}\{ x_t \in M(\delta) \} \geq 1 - \delta.
$$
Suppose that $R(\e, \delta)$ satisfies the following condition: for any positive $\e > 0$ and $\delta \in (0, 1)$ there exists a function $\hat{s} : \R^D \mapsto \R^D$ such that
\begin{align}
    \norm{\hat{s}}_{\calF_1} \leq R(\e, \delta), \quad \sup_{x \in M(\delta)} \norm{ \hat{s}(x) - \nabla \log p_t(x) } \leq \e. \label{eq:score_function_approx_general_condition}
\end{align}
Suppose there exists a $\delta \in (0, 1)$ satisfying:
\begin{align}
    R^4(\e/2, \delta) \cdot \delta \leq c_0 \e^4/\E\norm{x_t}^4, \quad \delta \leq c_1 (\e \sigma_t)^4 / D^2.
    \label{eq:approx_condition}
\end{align}
Above, both $c_0, c_1$ are universal positive constants.
Then, there exists an $\hat{s} : \R^D \mapsto \R^D$ such that:
\begin{align}
    \norm{\hat{s}}_{\calF_1} \leq R(\e/2, \delta), \quad \norm{\hat{s} - \nabla \log p_t}_{L_2(p_t)} \leq \e.
\end{align}
\end{myprop}
\begin{proof}
Let $\calE_G := \{ x_t \in M(\delta) \}$.
By assumption, we have that $\Pr(\calE_G) \geq 1-\delta$.
Put $s_\star := \nabla \log p_t$
and let $\hat{s} : \R^D \mapsto \R^D$ be as guaranteed by the assumption such that:
$$
    \norm{\hat{s}}_{\calF_1} \leq R(\e/2, \delta), \quad \sup_{x \in M(\delta)} \norm{\hat{s}(x) - s_\star(x)} \leq \e/2.
$$
Hence,
\begin{align*}
    \E_{x_t} \norm{\hat{s} - s_\star}^2 &= \E_{x_t} \norm{\hat{s} - s_\star}^2 \ind\{\calE_G\} + \E_{x_t} \norm{\hat{s} - s_\star}^2 \ind\{\calE_G^c\} \\
    &\leq \sup_{x \in M(\delta)} \norm{\hat{s}(x) - s_\star(x)}^2 + \sqrt{\E_{x_t}\norm{\hat{s}-s_\star}^4} \cdot \sqrt{\delta} \\
    &\leq (\e/2)^2 + \sqrt{\E_{x_t}\norm{\hat{s}-s_\star}^4} \cdot \sqrt{\delta} . 
\end{align*}
Consequently, by taking square root of both sides:
\begin{align*}
    \norm{\hat{s}-s_\star}_{L_2(p_t)} \leq \e/2 + \norm{\hat{s}-s_\star}_{L_4(p_t)} \cdot \delta^{1/4}.
\end{align*}
Let us now control $\norm{\hat{s}-s_\star}_{L_4(p_t)}$.
By triangle inequality and $\norm{\hat{s}(x)} \leq \norm{\hat{s}}_{\calF_1} \norm{x}$ for all $x$:
\begin{align*}
    \norm{\hat{s}-s_\star}_{L_4(p_t)} &\leq \norm{\hat{s}}_{L_4(p_t)} + \norm{s_\star}_{L_4(p_t)}  \\
    &\leq \norm{\hat{s}}_{\calF_1} \norm{x_t}_{L_4(p_t)} + \norm{s_\star}_{L_4(p_t)} \\
    &\leq R(\e/2, \delta) \norm{x_t}_{L_4(p_t)} + \norm{s_\star}_{L_4(p_t)} .
\end{align*}
To control $\norm{s_\star}_{L_4(p_t)}$, we use Tweedie's formula:
\begin{align*}
    \nabla \log p_t(x_t) = \E\left[\frac{m_t x_0 - x_t}{\sigma_t^2} \,\bigg|\, x_t \right] = -\frac{\E[w \mid x_t]}{\sigma_t}.
\end{align*}
Hence by Jensen's inequality and the tower property,
\begin{align*}
    \norm{s_\star}_{L_4(p_t)}^4 = \E \norm{ \nabla \log p_t(x_t) }^4 = \frac{1}{\sigma_t^4} \E\norm{ \E[w \mid x_t] }^4 \leq \frac{1}{\sigma_t^4} \E \norm{w}^4 \leq \frac{3 D^2}{\sigma_t^4}.
\end{align*}
Combining these calculations,
\begin{align*}
    \norm{\hat{s}-s_\star}_{L_4(p_t)} &\leq R(\e/2, \delta) \norm{x_t}_{L_4(p_t)} + \frac{3^{1/4}\sqrt{D}}{\sigma_t}.
\end{align*}
Hence,
\begin{align*}
    \norm{\hat{s}-s_\star}_{L_2(p_t)} \leq \e/2 + \left[R(\e/2, \delta)\norm{x_t}_{L_4(p_t)} + \frac{3^{1/4}\sqrt{D}}{\sigma_t}  \right] \cdot \delta^{1/4}
\end{align*}
Hence, if we set $\delta$ such that:
\begin{align*}
    R(\e/2, \delta) \norm{x_t}_{L_4(p_t)} \cdot \delta^{1/4} \leq \e/4, \quad \frac{3^{1/4}\sqrt{D}}{\sigma_t} \cdot \delta^{1/4} \leq \e/4,
\end{align*}
then we conclude that $\norm{\hat{s}-s_\star}_{L_2(p_t)} \leq \e$.
\end{proof}

We next turn to our main $L_\infty$ approximation result for Lipschitz functions.
We proceed in two steps.
First, we extend \Cref{lemma:bach_lipschitz_approx}
to vector-valued Lipschitz functions in a straightforward way. Then, we use the log dominance rule
to invert the result.
For the first step, we have the following result.
\begin{myprop}
\label{corollary:bach_lipschitz_approx_vector_valued}
Let $f : \R^d \mapsto \R^d$.
Suppose that $f$ is $L$-Lipschitz and $B$-bounded on $B_2(d, M)$. Set $K_d := d \cdot (B \vee LM)$.
For any $\gamma \geq O_d(1) \cdot K_d$, there exists an $f_\gamma \in \calF_1$ such that $\norm{f_\gamma}_{\calF_1} \leq  \gamma$ and:
\begin{align}
    \sup_{x \in B_2(d, M)} \norm{f(x)-f_\gamma(x)} \leq O_d(1) K_d \left(\frac{K_d}{\gamma}\right)^{2/(d+1)} \log\left(\frac{\gamma}{K_d}\right). \label{eq:bach_lipschitz_approx_vector_valued}
\end{align}
\end{myprop}
\begin{proof}
For $i \in [d]$, let $f_i(x) := \ip{e_i}{f(x)}$, where $e_i \in \R^d$ is the $i$-th standard basis vector.
We will apply \Cref{lemma:bach_lipschitz_approx} to each of the $f_i$'s.
Note that each $f_i$ is also $L$-Lipschitz and $B$-bounded on $B_2(d, M)$.
Hence, for every $i \in [d]$ there exists an $f_{\gamma,i} \in \calF_1$ with
$\norm{f_{\gamma,i}}_{\calF_1} \leq \gamma/d$ and:
$$
    \sup_{x \in B_2(d, M)} \abs{f_i(x) - f_{\gamma,i}(x)} \leq O_d(1) \frac{K_d}{d} \left(\frac{K_d}{\gamma}\right)^{2/(d+1)} \log\left(\frac{\gamma}{K_d}\right) =: \zeta.
$$
Choosing $f_\gamma := (f_{\gamma,1}, \dots, f_{\gamma,d})$ yields:
\begin{align*}
    \sup_{x \in B_2(d, M)} \norm{ f(x) - f_\gamma(x) } &= \sup_{x \in B_2(d, M)} \sqrt{\sum_{i=1}^{d} \abs{f_i(x) - f_{\gamma,i}(x)}^2 } \\
    &\leq \sqrt{ \sum_{i=1}^{d} \sup_{x\in B_2(d, M)} \abs{f_i(x) - f_{\gamma,i}(x)}^2 } \leq \sqrt{d} \zeta.
\end{align*}
To finish the claim, we bound the $\calF_1$ norm of $f_\gamma$. Since
\begin{align*}
    f_\gamma(x) = \sum_{i=1}^{d} e_i f_{\gamma,i}(x),
\end{align*}
by triangle inequality $\norm{f_\gamma}_{\calF_1} \leq \sum_{i=1}^{d} \norm{f_{\gamma,i}}_{\calF_1} \leq d \cdot  (\gamma/d) = \gamma$.
\end{proof}

We now execute the second step, 
where we invert the RHS of \Cref{corollary:bach_lipschitz_approx_vector_valued}
and solve for $\gamma$.
\lipschitzapprox*
\begin{proof}
Setting the RHS of \eqref{eq:bach_lipschitz_approx_vector_valued}
from \Cref{corollary:bach_lipschitz_approx_vector_valued} to $\e$ and rearranging terms, 
we need the following condition to hold (recall $K_d := d \cdot K$):
\begin{align*}
    \frac{\gamma}{K_d} \geq \left(\frac{O_d(1)K_d}{\e}\right)^{(d+1)/2} \log^{(d+1)/2}\left(\frac{\gamma}{K_d}\right).
\end{align*}
Using \Cref{prop:log_dominance_rule},
a sufficient condition is:
\begin{align*}
    \frac{\gamma}{K_d} \geq O_d(1) \left(\frac{O_d(1) K_d}{\e}\right)^{(d+1)/2} \log^{(d+1)/2}\left( O_d(1) \left(\frac{O_d(1) K_d}{\e}\right)^{(d+1)/2}\right).
\end{align*}
The claim now follows by simplifying these expressions
with our assumptions.
\end{proof}

\section{Uniform convergence for the DSM loss}
\label{sec:appendix:uniform_convergence}

Our first step is to establish the claimed
basic inequality \eqref{eq:DSM_basic_inequality}.
\begin{myprop}[Basic generalization inequality]
\label{prop:basic_generalization_bound}
Let $\scrF$ be any set of functions mapping $\R^D \mapsto \R^D$. 
Let $\hat{f}_t \in \scrF$ denote the DSM empirical risk minimizer:
\begin{align*}
    \hat{f}_t \in \arg\min_{f \in \scrF} \hat{\calL}_t(f).
\end{align*}
Then, we have for any $\e \geq 0$:
\begin{align*}
    \E_{\calD_t}[\calR_t(\hat{f}_t)] &\leq (1+\e)\inf_{f \in \scrF} \calR_t(f) + \E_{\calD_t} \sup_{f \in \scrF} [ \calL_t(f) - (1+\e) \hat{\calL}_t(f) ]+ \e \cdot C_t.
\end{align*}
\end{myprop}
\begin{proof}
For any $\e \geq 0$ and any $f \in \scrF$,
\begin{align*}
    \E_{\calD_t}[ \calR_t(\hat{f}_t) ] &= \E_{\calD_t}[ \calL_t(\hat{f}_t) - C_t ] &&\text{using \Cref{fact:dsm_score_matching_loss}} \\
    &= \E_{\calD_t}[ \calL_t(\hat{f}_t) - (1+\e) \hat{\calL}_t(\hat{f}_t) + (1+\e) \hat{\calL}_t(\hat{f}_t) - C_t ] \\
    &\leq \E_{\calD_t}[ \calL_t(\hat{f}_t) - (1+\e) \hat{\calL}_t(\hat{f}_t) + (1+\e) \hat{\calL}_t(f) - C_t ] &&\text{since $\hat{f}_t$ is an ERM} \\
    &= \E_{\calD_t}[ \calL_t(\hat{f}_t) - (1+\e) \hat{\calL}_t(\hat{f}_t) ] + (1+\e) \calL_t(f) - C_t && \text{since $\E_{\calD_t}[ \hat{\calL}_t(f) ] = \calL_t(f)$} \\
    &= \E_{\calD_t}[ \calL_t(\hat{f}_t) - (1+\e) \hat{\calL}_t(\hat{f}_t) ] + (1+\e) \calR_t(f) + \e  \cdot C_t &&\text{using \Cref{fact:dsm_score_matching_loss}} \\
    &\leq \E_{\calD_t}\sup_{f \in \scrF} [ \calL_t(f) - (1+\e) \hat{\calL}_t(f) ] + (1+\e) \calR_t(f) + \e  \cdot C_t.
\end{align*}
The claim now follows by taking the infimum of the RHS over $f \in \scrF$.
\end{proof}

The rest of this section will focus 
on the uniform convergence term 
in the basic inequality \eqref{eq:DSM_basic_inequality}.
We first define some notation which we will use in our analysis.
Let $\nu_\delta(\check{x}_0, \check{x}_t)$ denote a distribution over
pairs of \emph{truncated} vectors, parameterized by $\delta \in (0, 1)$, defined as follows:
\begin{align}
    \nu_\delta := \mathrm{Law}((x_0, x_t) \cdot \ind\{ \calE_x(\delta) \}), \quad \Pr\{ \calE_x(\delta) \} \geq 1-\delta. \label{eq:x_t_check_def}
\end{align}
Note that in the above definition, the event $\calE_x(\delta)$ lives in the \emph{joint} probability space of $(x_0, x_t)$.
The specifics of the event $\calE_x(\delta)$
are left unspecified for now, as they depend on the
underlying details of our latent structure.
However, we will require the following properties to hold almost surely
for some $\check{\mu}_{t,x}(\delta)$ and $\check{\mu}_{t,q}(\delta)$:
\begin{align}
    (\check{x}_0, \check{x}_t) \sim \nu_\delta \Longrightarrow \norm{\check{x}_t} \leq \check{\mu}_{t,x}(\delta) \textrm{  and  } \norm{\nabla \log q_t(\check{x}_t \mid \check{x}_0)} \leq \check{\mu}_{t,q}(\delta). \label{eq:x_t_check_bounds}
\end{align}
Next, define the population denoising loss over $\nu_\delta$ as:
\begin{align}
    \check{\calL}_t(f; \delta) := \E_{(\check{x}_0, \check{x}_t) \sim \nu_\delta} \norm{ f(\check{x}_t) - \nabla \log q_t(\check{x}_t \mid \check{x}_0) }^2. \label{eq:population_DSM_truncated}
\end{align}
Furthermore, 
given a dataset $\bar{\calD}_t = \{ (\bar{x}_0^i, \bar{x}_t^i) \}_{i=1}^{n}$,
the \emph{generalized empirical loss} is defined as
\begin{align}
    \hat{\calL}_t(f; \bar{\calD}_t) := \frac{1}{n}\sum_{i=1}^{n} \norm{f(\bar{x}_t^i) - \nabla \log q_t(\bar{x}_t^i \mid \bar{x}_0^i)}^2. \label{eq:empirical_DSM_general}
\end{align}
Note that the above definitions are used only in our truncation argument, and do not appear in the actual learning procedure.

The main result of this section is the following bound
on the uniform convergence term.
\begin{mylemma}
\label{lemma:fast_rate}
For $R_t \geq 1$, define
$\scrF_t := \{ s : \R^D \mapsto \R^D \mid \norm{s}_{\calF_1} \leq R_t \}$.
For $\e \in (0, 1]$, we have:
\begin{align*}
    &\E \sup_{f \in \scrF_t}[ \calL_t(f) - (1+\e) \hat{\calL}_t(f) ] \\
    &\leq \tilde{O}(1) (1+\e^{-1})\left[ \frac{R_t^2 \check{\mu}_{t,x}^2(n^{-5}) D + \check{\mu}_{t,q}^2(n^{-5})}{n} \right] + O(1)\frac{R_t^2 \norm{x_t}^2_{L_4(p_t)} + D / \sigma_t^2}{n^2}.
\end{align*}
\end{mylemma}
The proof of \Cref{lemma:fast_rate} follows immediately
from the following two results (invoking them both with $\delta=n^{-4}$).
The first result applies a truncation argument so that it suffices
to prove uniform convergence over truncated data.
\begin{myprop}
\label{prop:diffusion_fast_rate_intermediate}
Fix a $\delta \in (0, 1)$.
Define the truncated random pair $(\check{x}_0, \check{x}_t) \sim \nu_{\delta/n}$ (cf.~\eqref{eq:x_t_check_def}). 
Let the truncated dataset $\check{\calD}_t := \{(\check{x}_0^i, \check{x}_t^i)\}_{i=1}^{n}$ be $n$ iid copies of $(\check{x}_0, \check{x}_t)$, i.e., $\check{\calD}_t \sim \nu_{\delta/n}^{\otimes n}$.
For some $R_t \geq 1$, define $\scrF_t := \{ s : \R^D \mapsto \R^D \mid \norm{s}_{\calF_1} \leq R_t \}$.
For all $\e \in [0, 1]$, we have that:
\begin{align*}
    &\E\sup_{f \in \scrF_t}[ \calL_t(f) - (1+\e) \hat{\calL}_t(f) ] \\
    &\leq \E \sup_{f \in \scrF_t}[ \check{\calL}_t(f; \delta/n) - (1+\e) \hat{\calL}_t(f; \check{\calD}_t) ] + c(R_t^2 \norm{x_t}_{L_4(p_t)}^2 + \sigma_t^{-2} D) \cdot \delta^{1/2},
\end{align*}
where $c > 0$ is a universal constant.
\end{myprop}
\begin{proof}
Let $\calE_G$ denote the event $\calE_G := \{ \calD_t = \check{\calD}_t \}$.
By a union bound, $\Pr(\calE_G) \geq 1-\delta$.
Next we define for a dataset $\bar{\calD}_t$ the random variable:
\begin{align}
    \psi(\bar{\calD}_t) := \sup_{f \in \scrF_t} [ \calL_t(f) - (1+\e) \hat{\calL}_t(f; \bar{\calD}_t) ]. \label{eq:psi_def}
\end{align}
Applying \Cref{prop:random_variable_S_Scheck}:
\begin{align}
    \E[\psi(\calD_t)] \leq \E[\psi(\check{\calD}_n)] + ( \sqrt{\E[\psi^2(\calD_t)]} + \sqrt{\E[\psi^2(\check{\calD}_n)]}  ) \cdot \delta^{1/2}. \label{eq:fast_rate_truncation_step}
\end{align}
We next need to upper bound both:
\begin{align*}
    \E[ \psi^2 (\calD_t)], \quad \E[ \psi^2(\check{\calD}_n) ].
\end{align*}
To do this, we first derive a few intermediate bounds.
We start with:
\begin{align*}
    \calL_t(f) &= \E_{(x_0, x_t)}\norm{f(x_t) - \nabla \log q_t(x_t \mid x_0)}^2 \\
    &\leq 2 \E\norm{f(x_t)}^2 + 2 \E \norm{\nabla \log q_t(x_t \mid x_0)}^2 &&\text{since $(a+b)^2 \leq 2(a^2 + b^2)$} \\
    &\leq 2 R_t^2 \E\norm{x_t}^2 + 2 \E \norm{ (x_t-m_t x_0)/\sigma_t^2}^2 &&\text{since $\norm{f}_{\calF_1} \leq R_t$} \\
    &= 2 R_t^2 \E\norm{x_t}^2 + 2 \E \norm{ w/\sigma_t }^2 \\
    &= 2 R_t^2 \E \norm{x_t}^2 + 2 D/\sigma_t^2.
\end{align*}
Next, we have:
\begin{align*}
    \E \sup_{f \in \scrF_t} \hat{\calL}_t^2(f) &= \E \sup_{f \in \scrF_t} \left(\frac{1}{n}\sum_{i=1}^{n} \norm{f(x_t^i) - \nabla \log q_t(x_t^i \mid x_0^i)}^2\right)^2 \\
    &\leq \E\left(\frac{2R_t^2}{n}\sum_{i=1}^{n} \norm{x_t^i}^2 + \frac{2}{n\sigma_t^2}\sum_{i=1}^{n} \norm{w^i}^2\right)^2 &&\text{since $\norm{f}_{\calF_1} \leq R_t$} \\
    &\leq \frac{4R_t^4}{n} \sum_{i=1}^{n} \E\norm{x_t^i}^4 + \frac{4}{n\sigma_t^4}\sum_{i=1}^{n} \E\norm{w^i}^4 &&\text{Cauchy-Schwarz} \\
    &\leq 4 R_t^4 \E\norm{x_t}^4 + 12 D^2/\sigma_t^4.
\end{align*}
Hence,
\begin{align*}
    \E[ \psi^2(\calD_t) ] &\leq \E \sup_{f \in \scrF_t} [ \calL_t(f) - (1+\e) \hat{\calL}_t(f) ]^2 \\
    &\lesssim \sup_{f \in \scrF_t} \calL_t^2(f) + \E \sup_{f \in \scrF_t} \hat{\calL}^2_t(f) \\
    &\lesssim R_t^4 \E\norm{x_t}^4 + D^2/\sigma_t^4.
\end{align*}

Now we move on to bounding $\E[ \psi^2(\check{\calD}_n) ]$.
Defining $\calE_x := \calE_x(\delta/n)$,
\begin{align*}
    \check{x}_t - m_t \check{x}_0 = (x_t - m_t x_0) \cdot \ind\{ \calE_{x} \} = \sigma_t w \cdot \ind\{\calE_{x} \},
\end{align*}
and therefore:
\begin{align*}
    \E \norm{ ( \check{x}_t - m_t \check{x}_0)/\sigma_t^2 }^4 &= \E \norm{ w/\sigma_t}^4 \ind\{ \calE_x \} \leq \E \norm{w/\sigma_t^4} \leq 3D^2/\sigma_t^4.
\end{align*}
Hence:
\begin{align*}
    \E \sup_{f \in \scrF_t} \hat{\calL}_t^2(f; \check{\calD}_t) &= \E \sup_{f \in \scrF_t} \left(\frac{1}{n}\sum_{i=1}^{n} \norm{f(\check{x}_t^i) - \nabla \log q_t(\check{x}_t^i \mid \check{x}_0^i)}^2\right)^2 \\
    &\leq \E\left(\frac{2R_t^2}{n}\sum_{i=1}^{n} \norm{\check{x}_t^i}^2 + \frac{2}{n\sigma_t^2}\sum_{i=1}^{n} \norm{(\check{x}_t^i - m_t \check{x}_0^i)/\sigma_t^2}^2\right)^2 &&\text{since $\norm{f}_{\calF_1} \leq R_t$} \\
    &\leq \frac{4R_t^4}{n} \sum_{i=1}^{n} \E\norm{\check{x}_t^i}^4 + \frac{4}{n\sigma_t^4}\sum_{i=1}^{n} \E\norm{(\check{x}_t^i-m_t \check{x}_0^i)/\sigma_t^2}^4 &&\text{Cauchy-Schwarz} \\
    &\leq \frac{4R_t^4}{n} \sum_{i=1}^{n} \E\norm{{x}_t^i}^4 + \frac{4}{n\sigma_t^4}\sum_{i=1}^{n} \E\norm{({x}_t^i-m_t {x}_0^i)/\sigma_t^2}^4 \\
    &= \frac{4R_t^4}{n} \sum_{i=1}^{n} \E\norm{{x}_t^i}^4 + \frac{4}{n\sigma_t^4}\sum_{i=1}^{n} \E\norm{w^i/\sigma_t}^4 \\
    &\leq 4 R_t^4 \E\norm{x_t}^4 + 12 D^2/\sigma_t^4.
\end{align*}
Therefore, we conclude that:
\begin{align}
    \max\{ \E[\psi^2(\calD_t)], \E[\psi^2(\check{\calD}_t)] \} \lesssim R_t^4 \E\norm{x_t}^4 + \sigma_t^{-4} D^2. \label{eq:psi_squared_bounds}
\end{align}
Plugging \eqref{eq:psi_squared_bounds} into
\eqref{eq:fast_rate_truncation_step},
\begin{align*}
    \E[\psi(\calD_t)] &\leq \E[ \psi(\check{\calD}_t) ] + c(R_t^2 \norm{x_t}_{L_4(p_t)}^2 + \sigma_t^{-2} D) \cdot \delta^{1/2} \\
    &\leq \E \sup_{f \in \scrF_t} [ \calL_t(f) - (1+\e) \hat{\calL}_t(f; \check{\calD}_t) ] + c(R_t^2 \norm{x_t}_{L_4(p_t)}^2 + \sigma_t^{-2} D) \cdot \delta^{1/2} \\
    &\leq \E \sup_{f \in \scrF_t} [ \check{\calL}_t(f; \delta/n) - (1+\e) \hat{\calL}_t(f; \check{\calD}_t) ] + \sup_{f \in \scrF_t} [ \calL_t(f) - \check{\calL}_t(f; \delta/n) ] \\
    &\qquad+ c(R_t^2 \norm{x_t}_{L_4(p_t)}^2 + \sigma_t^{-2} D) \cdot \delta^{1/2}.
\end{align*}
Next, define:
\begin{align*}
    V_f((x, \bar{x})) := \norm{ f(\bar{x}) - \nabla \log q_t(\bar{x} \mid x) }^2.
\end{align*}
Observe by Jensen's inequality we can bound
\begin{align*}
    \E[ V_f^2((x_0, x_t)) ] \leq \E \sup_{f \in \scrF_t} \hat{\calL}_t^2(f) \lesssim R_t^4 \E\norm{x_t}^4 + D^2/\sigma_t^4.
\end{align*}
Therefore, by application of \Cref{prop:random_variable_S_Scheck},
\begin{align*}
    \E[ V_f((x_0, x_t)) ] &\leq \E[ V_f((\check{x}_0, \check{x}_t)) ] + \sqrt{ \E[ V_f^2((x_0, x_t)) ] } \cdot \sqrt{\delta/n} \\
    &\leq \E[ V_f((\check{x}_0, \check{x}_t)) ] + c' (R_t^2 \norm{x_t}_{L_4(p_t)}^2 +\sigma_t^{-2} D) \cdot \sqrt{\delta/n}.
\end{align*}
Hence,
\begin{align*}
     \sup_{f \in \scrF_t} [ \calL_t(f) - \check{\calL}_t(f; \delta/n) ] &= \sup_{f \in \scrF_t} [ \E[ V_f((x_0, x_t)) ] - \E[ V_f((\check{x}_0, \check{x}_t)) ]] \\
     &\leq c' (R_t^2 \norm{x_t}_{L_4(p_t)}^2+\sigma_t^{-2} D) \cdot \sqrt{\delta/n},
\end{align*}
from which the claim follows.
\end{proof}

The second result proves uniform convergence over truncated
inputs.
\begin{myprop}
\label{lemma:diffusion_fast_rate_intermediate_2}
Fix $\delta \in (0, 1)$.
Define the truncated random vectors
$(\check{x}_0, \check{x}_t) \sim \nu_{\delta/n}$ (cf.~\eqref{eq:x_t_check_def}),
and let $\check{D}_t \sim \nu_{\delta/n}^{\otimes n}$.
For some $R_t \geq 1$, define $\scrF_t := \{ s : \R^D \mapsto \R^D \mid \norm{s}_{\calF_1} \leq R_t \}$.
For all $\e \in (0, 1]$, we have that:
\begin{align*}
     &\E \sup_{f \in \scrF_t} [ \check{\calL}_t(f; \delta/n) - (1+\e) \hat{\calL}_t(f; \check{\calD}_t) ] \\
     &\leq \tilde{O}(1) (1+\e^{-1})\left[ \frac{R_t^2 \check{\mu}_{t,x}^2(\delta/n) D + \check{\mu}_{t,q}^2(\delta/n)}{n} \right] + O(1)\frac{R_t^2 \norm{x_t}^2_{L_4(p_t)} + D / \sigma_t^2}{n^2}.
\end{align*}
\end{myprop}
\begin{proof}
Let $\check{\mu}_{t,x} := \check{\mu}_{t,x}(\delta/n)$
and similarly $\check{\mu}_{t,q} := \check{\mu}_{t,q}(\delta/n)$.
We first observe that the following holds almost surely:
\begin{align*}
    \norm{ f(\check{x}_t) - \nabla \log q_t(\check{x}_t \mid \check{x}_0) } &\leq R_t \norm{\check{x}_t} + \norm{\nabla \log q_t(\check{x}_t \mid \check{x}_0) } \\
    &\leq R_t \check{\mu}_{t,x} + \check{\mu}_{t,q} 
    =: B_{\calH}.
\end{align*}
We consider the hypothesis of functions:
\begin{align*}
    \calH := \{ (x, \bar{x}) \mapsto \norm{f(\bar{x}) - \nabla \log q_t(\bar{x} \mid x)} \mid f \in \scrF_t \},
\end{align*}
defined over the support $\check{\calZ} := \mathrm{supp}((\check{x}_0, \check{x}_t))$, and
coupled with the loss function $\phi(z) = z^2$, which is $2$-smooth.
From~\citet[Theorem 1]{srebro2010fastrates}, 
we have with probability at least $1-\delta$ over $\check{D}_t$,
for all $h \in \calH$:\footnote{Note that we ignore the labels $y$ in the setup of \citet[Theorem 1]{srebro2010fastrates}, as they are immaterial.}
\begin{align}
    \E[ \phi(h(\check{x}_0, \check{x}_t)) ] \leq (1+\e) \frac{1}{n}\sum_{i=1}^{n} \phi(h(\check{x}_0^i, \check{x}_t^i)) + (1+\e^{-1}) c \left[ \log^3{n} \cdot \mathfrak{R}^2_n(\calH) + \frac{B^2_{\calH} \log(1/\delta)}{n} \right], \label{eq:srebro}
\end{align}
where $c > 0$ is a universal constant, and 
$\mathfrak{R}_n(\calH)$ denotes the Rademacher complexity of $\calH$:
\begin{align*}
    \mathfrak{R}_n(\calH) := \sup_{z_{1:n} \subset \check{\calZ} } \E_{\e} \sup_{h \in \calH} \frac{1}{n}\bigabs{ \sum_{i=1}^{n} h(z_i) \e_i  }.
\end{align*}
We now bound this Rademacher complexity term.
Let $\calG$ denote the shifted function class:
\begin{align*}
    \calG := \{ (x, \bar{x}) \mapsto f(\bar{x}) - \nabla \log q_t(\bar{x} \mid x) \mid f \in \scrF_t \}.
\end{align*}
Letting $g_0 \in \calG$ and $z_i = (x_i, \bar{x}_i) \in \check{Z}$ for $i \in [n]$ be arbitrary, we have:
\begin{align*}
    &\E_{\e} \sup_{h \in \calH} \bigabs{ \sum_{i=1}^{n} h(z_i) \e_i  } \\
    &= \E_{\e} \sup_{g \in \calG} \bigabs{ \sum_{i=1}^{n} \e_i\norm{g(z_i)} } \\
    &\leq \E_{\e} \sup_{g \in \calG} \bigabs{ \sum_{i=1}^{n} \e_i(\norm{g(z_i)}-\norm{g_0(z_i)}) } + \E \bigabs{ \sum_{i=1}^{n} \e_i \norm{g_0(z_i)} } \\
    &\leq \E_{\e} \sup_{g \in \calG} \bigabs{ \sum_{i=1}^{n} \e_i(\norm{g(z_i)}-\norm{g_0(z_i)}) } + \sqrt{ \sum_{i=1}^{n} \norm{g_0(z_i)}^2 } &&\text{Jensen's inequality} \\
    &\leq \E_{\e} \sup_{g, g' \in \calG} \bigabs{ \sum_{i=1}^{n} \e_i(\norm{g(z_i)}-\norm{g'(z_i)}) } + \sqrt{ \sum_{i=1}^{n} \norm{g_0(z_i)}^2 } &&\text{since $g_0 \in \calG$} \\
    &= \E_{\e} \sup_{g, g' \in \calG} \sum_{i=1}^{n} \e_i(\norm{g(z_i)}-\norm{g'(z_i)}) + \sqrt{ \sum_{i=1}^{n} \norm{g_0(z_i)}^2 } &&\text{since $\calG - \calG$ is symmetric} \\
    &\leq 2 \E_{\e} \sup_{g \in \calG} \sum_{i=1}^{n} \e_i \norm{g(z_i)} + \sqrt{n} B_{\calH} .
\end{align*}
Next, we proceed with \citet[Corollary 4]{maurer2016vector},
which allows to bound, for Rademacher random \emph{vectors} $\gamma_i \in \{\pm 1\}^{D}$,
\begin{align*}
    \E_{\e} \sup_{g \in \calG} \sum_{i=1}^{n} \e_i \norm{g(z_i)} &\leq \sqrt{2} \E_{\gamma} \sup_{g \in \calG} \sum_{i=1}^{n} \ip{\gamma_i}{g(z_i)} \\
    &= \sqrt{2} \E_{\gamma} \sup_{f \in \scrF_t} \sum_{i=1}^{n} \ip{\gamma_i}{f(\bar{x_i})} \\
    &\lesssim R_t \check{\mu}_{t,x} \sqrt{D n},
\end{align*}
where the last inequality uses \Cref{lemma:rademacher_complexity_F1}.
Putting the terms together,
\begin{align*}
    \mathfrak{R}_n(\calH) \lesssim ( R_t \check{\mu}_{t,x} \sqrt{D} + B_{\calH} ) \frac{1}{\sqrt{n}} \lesssim ( R_t \check{\mu}_{t,x} \sqrt{D} + \check{\mu}_{t,q} ) \frac{1}{\sqrt{n}}.
\end{align*}

From \eqref{eq:srebro},
with probability at least $1-1/n^4$,
for all $h \in \calH$:
\begin{align*}
    \E[ \phi(h(\check{x}_0, \check{x}_t)) ] &\leq (1+\e) \frac{1}{n}\sum_{i=1}^{n} \phi(h(\check{x}_0^i, \check{x}_t^i)) + (1+\e^{-1}) c \log^3{n} \left[ \frac{R_t^2 \check{\mu}_{t,x}^2 D + \check{\mu}_{t,q}^2}{n}  \right].
\end{align*}
That is, with probability at least $1-1/n^4$,
\begin{align*}
    \sup_{f \in \scrF_t} [ \check{\calL}_t(f; \delta/n) - (1+\e) \hat{\calL}_t(f; \check{\calD}_t) ] \leq \tilde{O}(1) (1+\e^{-1})\left[ \frac{R_t^2 \check{\mu}_{t,x}^2 D + \check{\mu}_{t,q}^2}{n} \right].
\end{align*}
Call this event $\calE'$.
We have that:
\begin{align*}
    &\E \sup_{f \in \scrF_t} [ \check{\calL}_t(f; \delta/n) - (1+\e) \hat{\calL}_t(f; \check{\calD}_t) ] \\
    &=\E\sup_{f \in \scrF_t} [ \check{\calL}_t(f; \delta/n) - (1+\e) \hat{\calL}_t(f; \check{\calD}_t) ] \ind\{\calE'\} + \E\sup_{f \in \scrF_t} [ \check{\calL}_t(f; \delta/n) - (1+\e) \hat{\calL}_t(f; \check{\calD}_t) ] \ind\{(\calE')^c\} \\
    &\lesssim \tilde{O}(1) (1+\e^{-1})\left[ \frac{R_t^2 \check{\mu}_{t,x}^2 D + \check{\mu}_{t,q}^2}{n} \right] + \frac{1}{n^2} \sqrt{\E \sup_{f \in \scrF_t} [ \check{\calL}_t(f; \delta/n) - (1+\e) \hat{\calL}_t(f; \check{\calD}_t) ]^2 }.
\end{align*}
To finish the proof, we observe that:
\begin{align*}
    \check{\calL}_t(f; \delta/n) &= \E_{(\check{x}_0, \check{x}_t) \sim \nu_{\delta/n}} \norm{ f(\check{x}_t) - \nabla \log q_t(\check{x}_t \mid \check{x}_0 )}^2 \\
    &\leq 2 \E\norm{ f(\check{x}_t) }^2 + 2 \E \norm{\nabla \log q_t(\check{x}_t \mid \check{x}_0) }^2 \\
    &\leq 2 R_t^2 \E \norm{\check{x}_t}^2 + 2 \E \norm{ (\check{x}_t - m_t \check{x}_0)/\sigma_t^2 }^2 \\
    &\leq 2 R_t^2 \E \norm{x_t}^2 + 2 \E\norm{ (x_t - m_t x_0)/\sigma_t^2 }^2 \\
    &= 2 R_t^2 \E \norm{x_t}^2 + 2D/\sigma_t^2.
\end{align*}
On the other hand, from \eqref{eq:psi_squared_bounds},
\begin{align*}
    \E \sup_{f \in \scrF_t} \check{\calL}_t^2(f; \check{\calD}_t ) \lesssim R_t^4 \E\norm{x_t}^4 + \sigma_t^{-4} D^2.
\end{align*}
Hence, combining these bounds together,
\begin{align*}
    &\E \sup_{f \in \scrF_t} [ \check{\calL}_t(f; \delta/n) - (1+\e) \hat{\calL}_t(f; \check{\calD}_t) ] \\
    &\lesssim\tilde{O}(1) (1+\e^{-1})\left[ \frac{R_t^2 \check{\mu}_{t,x}^2 D + \check{\mu}_{t,q}^2}{n} \right] + \frac{R_t^2 \norm{x_t}^2_{L_4(p_t)} + D / \sigma_t^2}{n^2}.
\end{align*}
\end{proof}

\section{Analysis of subspace structure (\Cref{sec:results:subspace})}
\label{sec:appendix:subspace}

We now specialize the previous approximation and uniform convergence results to the subspace structure setting.

\begin{myprop}
\label{prop:subspace_approx_bounded}
Fix an $M \geq 1$. 
For any $\e \in (0, \bar{L}M/2)$, there exists an $f_\e : \R^d \mapsto \R^d$ such that
$\sup_{z \in B_2(d, M)} \norm{f_\e(z) - \nabla \log \pi_t(z)} \leq \e$,
and 
\begin{align}
    \norm{f_\e}_{\calF_1} \leq \Rlin(\e, M) := O_d(1) (\bar{L} M)^{(d+3)/2} \e^{-(d+1)/2} \log^{(d+1)/2}( \bar{L}M/\e ) .
\end{align}
\end{myprop}
\begin{proof}
We will invoke \Cref{lemma:F1_lipschitz_approx}. 
To do this, we first observe for any $z \in B_2(d, M)$,
using \Cref{prop:linear_Linf_bound},
\begin{align*}
    \norm{\nabla \log \pi_t(z)} \leq \bar{L}(1 + \norm{z}) \leq 2\bar{L} M.
\end{align*}
On the other hand, we know that $\nabla \log \pi_t$ is $L$-Lipschitz.
The claim now follows from \Cref{lemma:F1_lipschitz_approx}.
\end{proof}

Our next task is to upgrade the previous result to an approximation result for the ambient score $\nabla \log p_t$, using \Cref{prop:score_function_approx_set_version}.
\begin{myprop}
\label{prop:subspace_approx}
Fix an $\e \in (0, 1)$.
There exists an $\hat{s} : \R^D \mapsto \R^D$ such that:
\begin{align}
    \norm{\hat{s}}_{\calF_1} \leq \tilde{O}_d(1) (\bar{L} (\mu_{t,z} \vee \beta))^{(d+3)/2} \e^{-(d+1)/2} + 2(D-d)/\sigma_t^2, \quad \norm{\hat{s} - \nabla \log p_t}_{L_2(p_t)} \leq \e.
\end{align}
\end{myprop}
\begin{proof}
Define
\begin{align*}
    M(\delta) := \left\{ \norm{U^\T x_t} \leq A_\delta \right\}, \quad A_\delta := c_0( \mu_{t,z} + \beta \sqrt{\log(1/\delta)} ).
\end{align*} 
We note that the condition
$\Pr\{ x_t \in M(\delta) \} \geq 1-\delta$ holds
for an appropriate choice of $c_0$.

Now, given $\e,\delta \in (0, 1)$,
from \Cref{prop:subspace_approx_bounded}
there exists $\hat{h} : \R^d \mapsto \R^d$ such that:
$$
    \norm{\hat{h}}_{\calF_1} \leq \Rlin(\e, A_\delta), \quad \sup_{z \in B_2(d, A_\delta)} \norm{\hat{h}(z) - \nabla \log \pi_t(z)} \leq \e.
$$
Embed $\hat{h}$ to a function $\hat{s} : \R^D \mapsto \R^D$ by:
$$
    \hat{s}(x) = U \hat{h}(U^\T x) - \frac{1}{\sigma_t^2} (I-UU^\T)x,
$$
and observe that (cf. \Cref{prop:subspace_structure_score_function}):
\begin{align*}
    \sup_{x \in M(\delta)} \norm{ \hat{s}(x) - s_\star(x) } &= \sup_{x \in M(\delta)} \norm{ U \hat{h}(U^\T x) - U \nabla \log \pi_t(U^\T x)} \\
    &\leq \sup_{x \in M(\delta)} \norm{ \hat{h}(U^\T x) - \nabla \log \pi_t(U^\T x)} \\
    &\leq \sup_{z \in B_2(d, A_\delta)} \norm{ \hat{h}(z) - \nabla \log \pi_t(z) } \\
    &\leq \e.
\end{align*}
Next, we bound the $\calF_1$-norm of $\hat{s}$.
To do this, we first bound the $\calF_1$-norm of 
$x \mapsto (I-UU^\T) x$ by
representing it by the following
sum of Dirac masses
\[
\sum_{i=1}^{D-d}\delta_{(u_{i},u_{i})}+\delta_{(-u_{i},-u_{i})},
\]
and hence
$\norm{x \mapsto (I-UU^\T) x}_{\calF_1} \leq 2 (D-d)$.
Next, recall by \Cref{fact:F1_norm_equiv} that
$\norm{x \mapsto U \hat{h}(U^\T x)}_{\calF_1} = \norm{h}_{\calF_1}$.
Combining these results,
\begin{align}
    \norm{\hat{s}}_{\calF_1} &\leq \norm{x \mapsto U \hat{h}(U^\T x)}_{\calF_1} + \frac{1}{\sigma_t^2} \norm{ x \mapsto (I-UU^\T) x }_{\calF_1} \nonumber \\
    &\leq \norm{\hat{h}}_{\calF_1} + \frac{2}{\sigma_t^2}(D-d) \leq \Rlin(\e, A_\delta) +  \frac{2}{\sigma_t^2}(D-d) =: R(\e, \delta). \label{eq:R_subspace_score}
\end{align}
That is, we have shown that 
for $\e > 0$ and $\delta \in (0, 1)$,
there exists a $\hat{s} : \R^D \mapsto \R^D$ such that:
\begin{align*}
    \norm{\hat{s}}_{\calF_1} \leq R(\e, \delta), \quad \sup_{x \in M(\delta)} \norm{\hat{s}(x) - \nabla \log p_t(x)} \leq \e.
\end{align*}
This verifies condition \eqref{eq:score_function_approx_general_condition} of \Cref{prop:score_function_approx_set_version}.
We now need to solve for a $\delta_\star$ which satisfies the conditions listed in \eqref{eq:approx_condition}.
By several applications of \Cref{prop:log_dominance_rule},
the conditions listed \eqref{eq:approx_condition} are
satisfied with a $\delta_\star \in (0, 1)$ satisfying:
$$
    \log(1/\delta_\star) \leq O_d(1) \log\left( \frac{\bar{L} D \mu_{t,x} \beta}{\e \sigma_t} \right).
$$
Since we do not track the exact form of the leading $O_d(1)$ constant, we skip the specific calculations.
The result now follows from
\Cref{prop:score_function_approx_set_version} after estimating $R(\e/2, \delta_\star)$. First, we bound,
\begin{align*}
    A_{\delta_\star} = c_0(\mu_{t,z} + \beta \sqrt{\log(1/\delta_\star)}) = \tilde{O}_d(1) (\mu_{t,z} \vee \beta).
\end{align*}
Hence,
\begin{align*}
    R(\e/2, \delta_\star) \leq \tilde{O}_d(1) (\bar{L} (\mu_{t,z} \vee \beta))^{(d+3)/2} \e^{-(d+1)/2}.
\end{align*}
Therefore, the result follows.
\end{proof}

Now we have the tools in place to prove \Cref{thm:score_estimation_latent},
our main score estimation result for this section.
\scoreestimationlatent*
\begin{proof}
First, by \Cref{prop:subspace_approx},
we know if we set $R_t$ to be
$$
    R_t = \tilde{O}_d(1) (\bar{L} (\mu_{t,z} \vee \beta))^{(d+3)/2} \e^{-(d+1)/2} + 2(D-d)/\sigma_t^2,
$$
then, we have $\inf_{s \in \scrF_t} \calL_t(s) \leq \e^2$.

Now we need to apply \Cref{lemma:fast_rate}.
To do this, we need to define our auxiliary truncated
random vectors (cf.~\eqref{eq:x_t_check_def}).
We choose the definition:
\begin{align}
    \calE_{x}(\delta) := \{ \norm{z_0} \leq \mu_0 + \beta \sqrt{2\log(2/\delta)}, \,\, \norm{w} \leq \sqrt{D} + \sqrt{2 \log(2/\delta)} \},
\end{align}
which by sub-Gaussian concentration followed by a union bound satisfies $\Pr\{ \calE_x(\delta) \} \geq 1 - \delta$.
Note that under this definition of $\calE_{x}(\delta)$, we can take:
\begin{align}
    \check{\mu}_{t,x}(\delta) \lesssim \mu_{t,x} + \beta \sqrt{\log(1/\delta)}, \quad \check{\mu}_{t,q}(\delta) \lesssim \sigma_t^{-1}(\sqrt{D} + \sqrt{\log(1/\delta)}).
\end{align}
By applying \Cref{lemma:fast_rate}, we obtain for $\gamma \in (0, 1)$,
\begin{align*}
    \E \sup_{f \in \scrF_t} [ \calL_t(f) - (1+\gamma)\hat{\calL}_t(f) ] \leq \tilde{O}(1) (1+\gamma^{-1})\frac{D}{n}\left[ R_t^2(\mu_{t,x} \vee \beta)^2 + 1/\sigma_t^2 \right].
\end{align*}
By the basic inequality \Cref{prop:basic_generalization_bound},
\begin{align*}
    \E_{\calD_t}[\calR_t(\hat{f}_t)] &\leq 2\e^2 + \tilde{O}_d(\gamma^{-1}) \frac{D}{n}(\bar{L} (\mu_{t,z} \vee \beta))^{d+3} \e^{-(d+1)} (\mu_{t,x} \vee \beta)^2 \\
    &\qquad+ \tilde{O}_d(\gamma^{-1}) \frac{D^2}{n \sigma_t^4}(\mu_{t,x} \vee \beta)^2 + \gamma \cdot C_t.
\end{align*}
We now optimize this expression over both $\e, \gamma \in (0,1)$.
We first optimize both expressions ignoring the constraint that $\e,\gamma < 1$.
First, optimizing over $\gamma$, we set
\begin{align*}
    \gamma = \tilde{O}_d(1) \sqrt{\frac{1}{C_t} \left[\frac{D}{n}(\bar{L} (\mu_{t,z} \vee \beta))^{d+3} \e^{-(d+1)} (\mu_{t,x} \vee \beta)^2 + \frac{D^2}{n \sigma_t^4}(\mu_{t,x} \vee \beta)^2 \right]},
\end{align*}
and from this we obtain:
\begin{align*}
    \E_{\calD_t}[\calR_t(\hat{f}_t)] &\leq 2\e^2 + \sqrt{ \tilde{O}_d(1) \frac{C_t D}{n}(\bar{L} (\mu_{t,z} \vee \beta))^{d+3} \e^{-(d+1)} (\mu_{t,x} \vee \beta)^2 } + \sqrt{ \tilde{O}_d(1) \frac{C_t D^2}{n \sigma_t^4}(\mu_{t,x} \vee \beta)^2 }.
\end{align*}
Now optimizing over $\e$, we set
\begin{align*}
    \e = \tilde{O}_d(1) \left[ \frac{C_t D}{n}(\bar{L} (\mu_{t,z} \vee \beta))^{d+3} (\mu_{t,x} \vee \beta)^2  \right]^{1/(d+5)},
\end{align*}
and obtain:
\begin{align*}
    \E_{\calD_t}[\calR_t(\hat{f}_t)] \leq \tilde{O}_d(1) \left[  \frac{C_t D}{n}(\bar{L} (\mu_{t,z} \vee \beta))^{d+3} (\mu_{t,x} \vee \beta)^2 \right]^{2/(d+5)} + \tilde{O}_d(1)  \sqrt{ \frac{C_t D^2}{n \sigma_t^4}(\mu_{t,x} \vee \beta)^2 }.
\end{align*}
The proof concludes by setting $n$ large enough so that both $\e, \gamma < 1$.
\end{proof}

We now restate and prove \Cref{corollary:sampling_subspace},
our main end-to-end bound for the latent subspace case.
\samplingsubspace*
\begin{proof}
Using the bounds
$\mu_{t,z} \leq \mu_z$ and 
$\mu_{t,x} \leq \mu_x$,
from \Cref{thm:score_estimation_latent}
we have that the following ERM bound holds for all $t \in [0, T]$:
\begin{align*}
    \E_{\calD_t}[\calR_t(\hat{f}_t)] \leq \tilde{O}_d(1) \left[  \frac{D^2}{n \sigma_t^2}(\bar{L} (\mu_{z} \vee \beta))^{d+3} (\mu_{x}^2 \vee \beta^2) \right]^{2/(d+5)} + \tilde{O}_d(1)  \sqrt{ \frac{D^3}{n \sigma_t^6}(\mu_{x}^2 \vee \beta^2) }. 
\end{align*}
Furthermore, since $e^{-x} \leq 1 - x/2$ for $x \in [0, 1.59]$,
then
for $t \leq 0.795$ we have 
$$
    \sigma_t^2 = 1-\exp(-2t) \geq t \Longrightarrow 1/\sigma_t^2 \leq 1/t.
$$
On the other hand, for
$t > 0.795$, we have the bound
\begin{align*}
    \sigma_t^2 = 1 - \exp(-2t) \geq 1-\exp(-1.59) \geq 0.796 \Longrightarrow 1/\sigma_t^2 \lesssim 1.
\end{align*}
Combining these inequalities we have that $1 / \sigma_t^2 \leq 1 / \sigma_{\zeta}^2 \lesssim 1/\zeta$ for all $t \geq \zeta$.

Hence, using the 
choice of $T$, $N$ from \eqref{eq:diffusion_T_N_setting} and $\{t_i\}_{i=0}^{N}$ as
specified in \Cref{thm:sampler_quality_from_L2_score},
we have that:
\begin{align*}
    &\sum_{k=0}^{N-1} \gamma_k \E_{p_{T-t_k}} \norm{ \hat{f}_{T-t_k} - \nabla \log p_{T-t_k}}^2 \\
    &\lesssim T \left[ \tilde{O}_d(1) \left[  \frac{D^2}{n \zeta}(\bar{L} (\mu_{z} \vee \beta))^{d+3} (\mu_{x}^2 \vee \beta^2) \right]^{2/(d+5)} + \tilde{O}_d(1)  \sqrt{ \frac{D^3}{n \zeta^3}(\mu_{x}^2 \vee \beta^2) } \right] \\
    &\lesssim \log\left(\frac{\sqrt{D} \vee \mu_0}{\e}\right) \left[ \tilde{O}_d(1) \left[  \frac{D^2}{n \zeta}(\bar{L} (\mu_{z} \vee \beta))^{d+3} (\mu_{x}^2 \vee \beta^2) \right]^{2/(d+5)} + \tilde{O}_d(1)  \sqrt{ \frac{D^3}{n \zeta^3}(\mu_{x}^2 \vee \beta^2) } \right].
\end{align*}
Hence, in order to make $\e_{\mathrm{score}}^2 \leq \e^2$, we need to take $n$ large enough such that
the following conditions hold:
\begin{align*}
    \log\left(\frac{\sqrt{D} \vee \mu_0}{\e}\right) \tilde{O}_d(1) \left[  \frac{D^2}{n \zeta}(\bar{L} (\mu_{z} \vee \beta))^{d+3} (\mu_{x}^2 \vee \beta^2) \right]^{2/(d+5)} &\lesssim \e^2, \\
     \log\left(\frac{\sqrt{D} \vee \mu_0}{\e}\right) \tilde{O}_d(1)  \sqrt{ \frac{D^3}{n \zeta^3}(\mu_{x}^2 \vee \beta^2) } &\lesssim \e^2.
\end{align*}
Hence, we need to take $n$ satisfying:
\begin{align*}
    n \geq \tilde{O}_d(1) \max\left\{  \frac{D^2}{\zeta}(\bar{L} (\mu_{z} \vee \beta))^{d+3} (\mu_{x}^2 \vee \beta^2) \cdot \e^{-(d+5)}, \frac{D^3}{\zeta^3} (\mu_x^2 \vee \beta^2) \cdot \e^{-4} \right\}.
\end{align*}
On the other hand we also need to take $n \geq n_0(\zeta)$
(cf.~\eqref{eq:ERM_score_matching_subspace_burnin}).
The claim now follows.
\end{proof}

\section{Analysis of independent components (\Cref{sec:results:independent_components})}
\label{sec:appendix:independent_components}

We follow a very similar structure as in \Cref{sec:appendix:subspace}.
We first start with an approximation result.

\begin{myprop}
\label{prop:ind_components_approx_bounded}
For an $M \geq 1$ and $i \in [K]$. For any $\e \in (0, \bar{L}_i M/2)$, there exists an $f_\e : \R^{d_i} \mapsto \R^{d_i}$ such that
$\sup_{z \in B_2(d_i, M)} \norm{f_\e(z) - \nabla \log \pi_t^{(i)}(z)} \leq \e$, and
\begin{align}
    \norm{f_\e}_{\calF_1} \leq \Rind^{(i)}(\e, M) := O_d(1) (\bar{L}_i M)^{(d+3)/2} \e^{-(d+1)/2} \log^{(d+1)/2}(\bar{L}_i M/\e).
\end{align}
\end{myprop}
\begin{proof}
The proof is nearly identical to \Cref{prop:subspace_approx_bounded},
and therefore we omit the details.
\end{proof}

Next, we upgrade the previous approximation result 
to approximation in $L_2(p_t)$.
\begin{myprop}
\label{prop:ind_components_approx}
Fix $\e_1, \dots, \e_K \in (0, 1)$.
There exists
an $\hat{s} : \R^{D} \mapsto \R^{D}$ satisfying:
\begin{align*}
    \norm{\hat{s}}_{\calF_1} \leq \sum_{i=1}^{K} \tilde{O}_{d_i}(1)(\bar{L}_i(\mu_{t,x}^{(i)} \vee \beta))^{(d_i+3)/2} \e_i^{-(d_i+1)/2}, \quad \norm{ \hat{s} - \nabla \log p_t }_{L_2(p_t)} \leq \sqrt{\sum_{i=1}^{K} \e_i^2}.
\end{align*}
\end{myprop}
\begin{proof}
Recall that $P_i \in \R^{d_i \times D}$ selects the coordinates
corresponding to the $i$-th variable group (cf.~\Cref{prop:independent_components_score_function}). 
Define the sets $M^{(i)}(\delta)$ as:
\begin{align*}
    M^{(i)}(\delta) := \left\{ x \in \R^{D} \mid \norm{ P_i U^\T x } \leq A_\delta^{(i)} \right\}, \quad A_\delta^{(i)} := c_0( \mu_{t,x}^{(i)} + \beta_i \sqrt{\log(1/\delta)} ), \quad i \in [K].
\end{align*}
With appropriate choice of $c_0$, we have that
$\Pr_{ x_t \sim p_t }\{ x_t \in M^{(i)}(\delta) \} \geq 1-\delta$.

Given $\e,\delta \in (0, 1)$,
from \Cref{prop:ind_components_approx_bounded},
there exists $\hat{h}_i : \R^d \mapsto \R^d$ for $i \in [K]$ such that:
$$
    \norm{\hat{h}_i}_{\calF_1} \leq \Rind^{(i)}(\e, A_\delta^{(i)}), \quad \sup_{z \in B_2(d, A_\delta^{(i)})} \norm{\hat{h}_i(z) - \nabla \log \pi_t^{(i)}(z)} \leq \e.
$$
Now define $\hat{s}_i := U P_i^\T \hat{h}_i(P_i U^\T x)$.
Observe that:
\begin{align*}
    \sup_{x \in M^{(i)}(\delta)} \norm{ \hat{s}_i(x) - U P_i^\T \nabla \log \pi_t^{(i)}(P_i U^\T x) } &\leq \sup_{x \in M^{(i)}(\delta)} \norm{ \hat{h}_i(P_i U^\T x) - \nabla \log \pi_t^{(i)}( P_i U^\T x) } \\
    &\leq \sup_{z \in B_2(d_i, A_\delta^{(i)})} \norm{ \hat{h}_i(z) - \nabla \log \pi_t^{(i)}(z) } \\
    &\leq \e.
\end{align*}
Next, observe that $\norm{\hat{s}_i}_{\calF_1} = \norm{\hat{h}_i}_{\calF_1}$ by \Cref{fact:F1_norm_equiv}.
Invoking \Cref{prop:score_function_approx_set_version}
as is done in the proof of \Cref{prop:subspace_approx},
we have that for all $i \in [K]$,
$\norm{ \hat{s}_i - U P_i^\T \nabla \log \pi_t^{(i)}(P_i U^\T \cdot) }_{L_2(p_t)} \leq \e_i$ and
\begin{align*}
    \norm{\hat{s}_i}_{\calF_1} \leq \tilde{O}_{d_i}(1) (\bar{L}_i (\mu_{t,z}^{(i)} \vee \beta_i))^{(d_i+3)/2} \e_i^{-(d+1)/2}.
\end{align*}
Recall by \Cref{prop:independent_components_score_function} we have:
\begin{align*}
    \nabla \log p_t(x) = \sum_{i=1}^{K} U P_i^\T \nabla \log \pi_t^{(i)}( P_i U^\T x ).
\end{align*}
Hence, setting $\hat{s} = \sum_{i=1}^{K} \hat{s}_i$, we have that
\begin{align*}
    &\norm{\hat{s} - \nabla \log p_t}_{L_2(p_t)}^2 \\
    &= \E \bignorm{\sum_{i=1}^{K} U P_i^\T (\hat{h}_i(P_i U^\T x_t) - \nabla \log \pi_t^{(i)}(P_i U^\T x_t))}^2 \\
    &= \sum_{i=1}^{K} \E \norm{U P_i^\T (\hat{h}_i(P_i U^\T x_t) - \nabla \log \pi_t^{(i)}(P_i U^\T x_t))}^2 &&\text{since $P_j P_i^\T = 0$ for $i \neq j$} \\
    &= \sum_{i=1}^{K} \norm{ \hat{s}_i - U P_i^\T \nabla \log \pi_t^{(i)}(P_i U^\T \cdot) }_{L_2(p_t)}^2 \\
    &\leq \sum_{i=1}^{K} \e_i^2.
\end{align*}
Furthermore,
\begin{align*}
    \norm{\hat{s}}_{\calF_1} \leq \sum_{i=1}^{K} \norm{\hat{s}_i}_{\calF_1} \leq \sum_{i=1}^{K} \tilde{O}_{d_i}(1) (\bar{L}_i (\mu_{t,z}^{(i)} \vee \beta_i))^{(d_i+3)/2} \e_i^{-(d+1)/2}.
\end{align*}
\end{proof}

We now prove \Cref{thm:score_estimation_ind_components}, our
score estimation result for the independent components setting.
\scoreestimationindcomponents*
\begin{proof}
Here we minic the proof of \Cref{thm:score_estimation_latent}.
First, by \Cref{prop:ind_components_approx},
we know if we set $R_t$ as
$$
    R_t = \sum_{i=1}^{K} \tilde{O}_{d_i}(1)(\bar{L}_i(\mu_{t,x}^{(i)} \vee \beta_i))^{(d_i+3)/2} \e_i^{-(d_i+1)/2},
$$
then, we have $\inf_{s \in \scrF_t} \calL_t(s) \leq \sum_{i=1}^{K} \e_i^2$.

Our next step is to apply \Cref{lemma:fast_rate}.
To do this we need to define auxiliary truncated random vectors
(cf.~\eqref{eq:x_t_check_def}). In this case, we use the definition:
\begin{align}
    \calE_{x}(\delta) := \bigcap_{i \in [K]} \{ \norm{z_0^{(i)}} \leq \mu_0^{(i)} + \beta_i \sqrt{2 \log(2K/\delta)} \} \cap \{ \norm{w} \leq \sqrt{D} + \sqrt{2\log(2/\delta)} \},
\end{align}
which satisfies $\Pr\{ \calE_{x}(\delta) \} \geq 1 - \delta$.
We have
\begin{align*}
    \check{\mu}_{t,x}(\delta) \lesssim \mu_{t,x} + \beta \sqrt{\log(K/\delta)}, \quad \check{\mu}_{t,q}(\delta) \lesssim \sigma_t^{-1} (\sqrt{D} + \sqrt{\log(1/\delta)}).
\end{align*}
By applying \Cref{lemma:fast_rate}, we obtain for $\gamma \in (0, 1)$,
\begin{align*}
    &\E \sup_{f \in \scrF_t} [ \calL_t(f) - (1+\gamma)\hat{\calL}_t(f) ] \\
    &\leq \tilde{O}(1) \gamma^{-1}\frac{D}{n}\left[ R_t^2(\mu_{t,x} \vee \beta)^2 + 1/\sigma_t^2 \right] \\
    &\leq \gamma^{-1}\frac{DK}{n} \sum_{i=1}^{K} \tilde{O}_{d_i}(1)(\bar{L}_i(\mu_{t,x}^{(i)} \vee \beta_i))^{d_i+3} \e_i^{-(d_i+1)}(\mu_{t,x} \vee \beta)^2 + \tilde{O}(1)\frac{D}{\gamma n \sigma_t^2}
\end{align*}
By the basic inequality \Cref{prop:basic_generalization_bound},
\begin{align*}
    \E_{\calD_t}[\calR_t(\hat{f}_t)] &\leq 2\sum_{i=1}^{K} \e_i^2 +  \gamma^{-1}\frac{DK}{n} \sum_{i=1}^{K} \tilde{O}_{d_i}(1)(\bar{L}_i(\mu_{t,x}^{(i)} \vee \beta_i))^{d_i+3} \e_i^{-(d_i+1)}(\mu_{t,x} \vee \beta)^2 \\
    &\qquad+ \tilde{O}(1)\frac{D}{\gamma n \sigma_t^2} + \gamma \cdot C_t.
\end{align*}
We now need to optimize over both $\e_i, \gamma \in (0, 1)$. 
We first set $\gamma$ as:
\begin{align*}
    \gamma = \sqrt{\frac{1}{C_t} \cdot \left[ \frac{DK}{n} \sum_{i=1}^{K} \tilde{O}_{d_i}(1)(\bar{L}_i(\mu_{t,x}^{(i)} \vee \beta_i))^{d_i+3} \e_i^{-(d_i+1)}(\mu_{t,x} \vee \beta)^2 +  \tilde{O}(1)\frac{D}{n \sigma_t^2} \right] },
\end{align*}
from which we obtain:
\begin{align*}
    \E_{\calD_t}[ \calR_t(\hat{f}_t) ] \leq 2 \sum_{i=1}^{K} \e_i^2 + \sum_{i=1}^{K} \sqrt{\tilde{O}_{d_i}(1)  \frac{C_t D K}{n} (\bar{L}_i(\mu_{t,x}^{(i)} \vee \beta_i))^{d_i+3} \e_i^{-(d_i+1)}(\mu_{t,x} \vee \beta)^2 } + \sqrt{ \tilde{O}(1) \frac{C_t D }{n \sigma_t^2} }.
\end{align*}
We now set $\e_i$ as:
\begin{align*}
    \e_i = \tilde{O}_{d_i}(1) \left[ \frac{C_t D K}{n} (\bar{L}_i(\mu_{t,x}^{(i)} \vee \beta_i))^{d_i+3} (\mu_{t,x} \vee \beta)^2 \right]^{1/(d_i+5)},
\end{align*}
and obtain:
\begin{align*}
    \E_{\calD_t}[\calR_t(\hat{f}_t)] \leq \sum_{i=1}^{K} \tilde{O}_{d_i}(1) \left[  \frac{C_t D K}{n}(\bar{L}_i (\mu_{t,x}^{(i)} \vee \beta_i))^{d_i+3} (\mu_{t,x} \vee \beta)^2 \right]^{2/(d_i+5)} + \sqrt{ \tilde{O}(1) \frac{C_t D}{n \sigma_t^2} }.
\end{align*}
The proof concludes by setting $n$ large enough so that all of $\e_i, \gamma < 1$.
\end{proof}

Finally, we conclude with \Cref{corollary:sampling_ind_components},
which provides an end-to-end sampling bound.
\samplingindcomponents*
\begin{proof}
We follow the proof of \Cref{corollary:sampling_subspace}.
Using the bounds
$\mu_{t,x}^{(i)} \leq \mu_x^{(i)}$ and 
$\mu_{t,x} \leq \mu_x$,
from \Cref{thm:score_estimation_ind_components}
we have that the following ERM bound holds for all $t \in [0, T]$:
\begin{align*}
    \E_{\calD_t}[\calR_t(\hat{s}_t)] \leq \sum_{i=1}^{K} \tilde{O}_{d_i}(1) \left[  \frac{D^2 K}{\sigma_t^2 n}(\bar{L}_i (\mu_{x}^{(i)} \vee \beta_i))^{d_i+3} (\mu_{x} \vee \beta)^2 \right]^{\frac{2}{d_i+5}} + \tilde{O}(1) \sqrt{ \frac{D^2}{\sigma_t^4 n}(\mu_{x} \vee \beta)^2 }.
\end{align*}
Recalling that $1/\sigma_t^{-2} \lesssim 1/\zeta$ for all $t \geq \zeta$, using the 
choice of $T$, $N$ from \eqref{eq:diffusion_T_N_setting} and $\{t_i\}_{i=0}^{N}$
we have that:
\begin{align*}
    &\sum_{k=0}^{N-1} \gamma_k \E_{p_{T-t_k}} \norm{ \hat{f}_{T-t_k} - \nabla \log p_{T-t_k}}^2 \\
    &\lesssim \log\left(\frac{\sqrt{D} \vee \mu_0}{\e}\right) 
     \sum_{i=1}^{K} \tilde{O}_{d_i}(1)  \left[ \frac{D^2 K}{\zeta n}(\bar{L}_i (\mu_{x}^{(i)} \vee \beta_i))^{d_i+3} (\mu_{x} \vee \beta)^2 \right]^{\frac{2}{d_i+5}} \\
    &\qquad + \log\left(\frac{\sqrt{D} \vee \mu_0}{\e}\right) \tilde{O}(1) \sqrt{ \frac{D^2}{\zeta^2 n}(\mu_{x} \vee \beta)^2 }.
\end{align*}
Hence, in order to make $\e_{\mathrm{score}}^2 \leq \e^2$, we need to take $n$ large enough such that
the following conditions hold:
\begin{align*}
    \log\left(\frac{\sqrt{D} \vee \mu_0}{\e}\right) 
     \sum_{i=1}^{K} \tilde{O}_{d_i}(1)  \left[ \frac{D^2 K}{\zeta n}(\bar{L}_i (\mu_{x}^{(i)} \vee \beta_i))^{d_i+3} (\mu_{x} \vee \beta)^2 \right]^{\frac{2}{d_i+5}} &\lesssim \e^2, \\
     \log\left(\frac{\sqrt{D} \vee \mu_0}{\e}\right) \tilde{O}(1) \sqrt{ \frac{D^2}{\zeta^2 n}(\mu_{x} \vee \beta)^2 } &\lesssim \e^2.
\end{align*}
Hence, we need to take $n$ satisfying:
\begin{align*}
    n \geq (\mu_x \vee \beta)^2 \max\left\{ \max_{i \in [K]} \left\{ \frac{\tilde{O}_{d_i}(1) D^2}{\zeta} K^{(d_i+7)/2} (\bar{L}_i (\mu_x^{(i)} \vee \beta))^{d_i+3} \cdot \e^{-(d_i+5)} \right\}, \frac{\tilde{O}(1) D^2}{\zeta^2}  \cdot \e^{-4}\right\} .
\end{align*}
On the other hand we also need to take $n \geq n_0(\zeta)$
(cf.~\eqref{eq:ERM_score_matching_ind_components_burnin}).
The claim now follows.
\end{proof}

\end{document}